\documentclass[conference]{IEEEtran}
\IEEEoverridecommandlockouts
\usepackage{cite}
\usepackage{amsmath,amssymb,amsfonts}
\usepackage{algorithm}
\usepackage{algorithmic}
\usepackage{graphicx}
\usepackage{textcomp}
\usepackage{xcolor}
\usepackage{booktabs}
\usepackage{multirow}
\usepackage{subcaption}
\usepackage{ulem}

\def\BibTeX{{\rm B\kern-.05em{\sc i\kern-.025em b}\kern-.08em
    T\kern-.1667em\lower.7ex\hbox{E}\kern-.125emX}}
\begin{document}

\title{Memory-enhanced Invariant Prompt Learning for Urban Flow Prediction under Distribution Shifts\\
}

\author{\IEEEauthorblockN{Haiyang Jiang}
\IEEEauthorblockA{
\textit{The University of Queensland}\\
Brisbane, Australia \\
haiyang.jiang@uq.edu.au}
\\
\IEEEauthorblockN{Nguyen Quoc Viet Hung}
\IEEEauthorblockA{
\textit{Griffith University}\\
Gold Coast, Australia \\
quocviethung1@gmail.com}
\and
\IEEEauthorblockN{Tong Chen}
\IEEEauthorblockA{
\textit{The University of Queensland}\\
Brisbane, Australia \\
tong.chen@uq.edu.au}
\\
\IEEEauthorblockN{Yuan Yuan,Yong Li}
\IEEEauthorblockA{
\textit{Tsinghua University}\\
Beijing, China \\
y-yuan20@mails.tsinghua.edu.cn \\ 
liyong07@tsinghua.edu.cn}
\and
\IEEEauthorblockN{Wentao Zhang}
\IEEEauthorblockA{
\textit{Peking University}\\
Beijing, China \\
wentao.zhang@pku.edu.cn}
\\
\IEEEauthorblockN{Lizhen Cui}
\IEEEauthorblockA{
\textit{Shandong University}\\
Jinan, China \\
clz@sdu.edu.cn}
}
\maketitle

\begin{abstract}
Urban flow prediction is a classic spatial-temporal forecasting task that estimates the amount of future traffic flow for a given location. Though models represented by Spatial-Temporal Graph Neural Networks (STGNNs) have established themselves as capable predictors, they tend to suffer from distribution shifts that are common with the urban flow data due to the dynamics and unpredictability of spatial-temporal events. To generalize STGNNs to out-of-distribution (OOD) data, a key strategy is to discover and disentangle the causal, invariant patterns from the ones that are variant and dependent on the spatial-temporal environments. However, to achieve this, existing OOD-robust methods heavily rely on learning and mimicking the underlying environments to facilitate changes in the data and consequently introduce different variant patterns. Unfortunately, in spatial-temporal applications, the dynamic environments can hardly be quantified via a fixed number of parameters, whereas learning time- and location-specific environments can quickly become computationally prohibitive. 
In this paper, we propose a novel framework named Memory-enhanced Invariant Prompt learning (MIP) for urban flow prediction under constant distribution shifts. Specifically, MIP is equipped with a learnable memory bank that is trained to memorize the causal features within the spatial-temporal graph. By querying a trainable memory bank that stores the causal features, we adaptively extract invariant and variant prompts (i.e., patterns) for a given location at every time step. Then, instead of intervening the raw data based on simulated environments, we directly perform intervention on variant prompts across space and time. With the intervened variant prompts in place, we use invariant learning to minimize the variance of predictions, so as to ensure that the predictions are only made with invariant features. With extensive comparative experiments on two public urban flow datasets, we thoroughly demonstrate the robustness of MIP against OOD data. 
\end{abstract}

\begin{IEEEkeywords}
Urban Flow Prediction, Distribution Shift, Invariant Prompt
\end{IEEEkeywords}

\section{Introduction}
Urban flow prediction, which involves forecasting traffic, pedestrian, and public transportation dynamics, is essential for efficient urban planning and management. 
Applications include smart city~\cite{giuffre2012novel,vijayalakshmi2021attention,9835423}, public transportation management~\cite{9835704,10184520,li2017diffusion,wu2019graph}, and ride-sharing services~\cite{chen2021spatial,zhang2017deep}. 
The insights from predictive models support both immediate traffic management and long-term infrastructure planning, contributing to smarter, more sustainable urban environments with improved mobility and resource allocation. 

\begin{figure}[tp]
    \centering
    \includegraphics[width=0.45\textwidth]{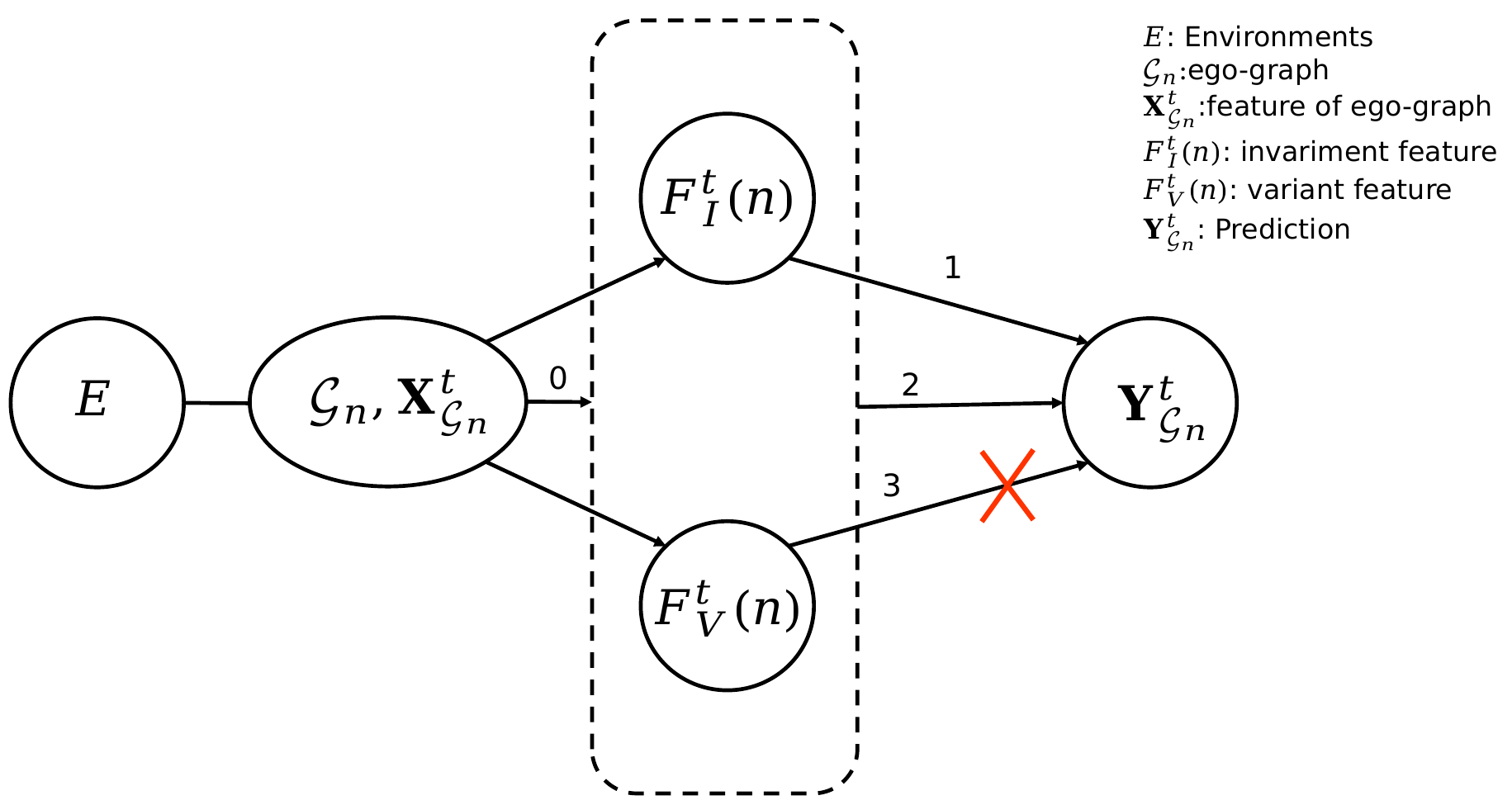}
    \caption{\small  Causal graph of spatial-temporal prediction. Previous STGNNs learn the embeddings containing both invariant and variant features and make predictions based on the relation line 2, but they are always misled to learn the spurious relation line 3 because of the variant features. In this study, we try to separate the invariant and variant features and make predictions based on the causal relationship, line 1.}
    \label{fig:scm}
    \vspace{-5mm}
\end{figure}


\begin{figure*}[htbp]
\centering
    \begin{subfigure}{0.3\linewidth}
        \centering
        \includegraphics[width=\textwidth]{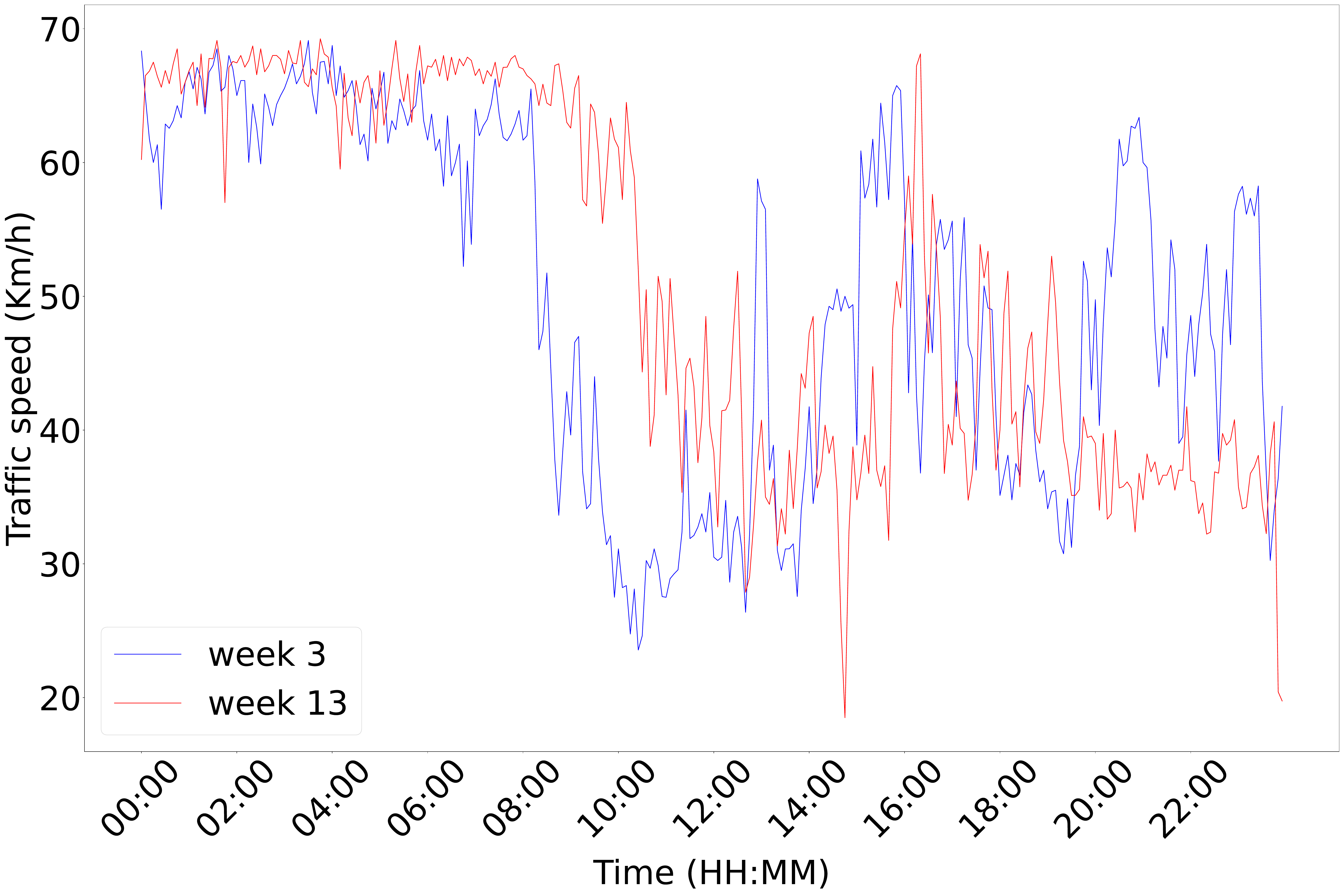}
        \caption{Traffic speed at node A.}
        \label{fig:OOD_1}
    \end{subfigure}%
    \begin{subfigure}{0.3\linewidth}
        \centering
        \includegraphics[width=\textwidth]{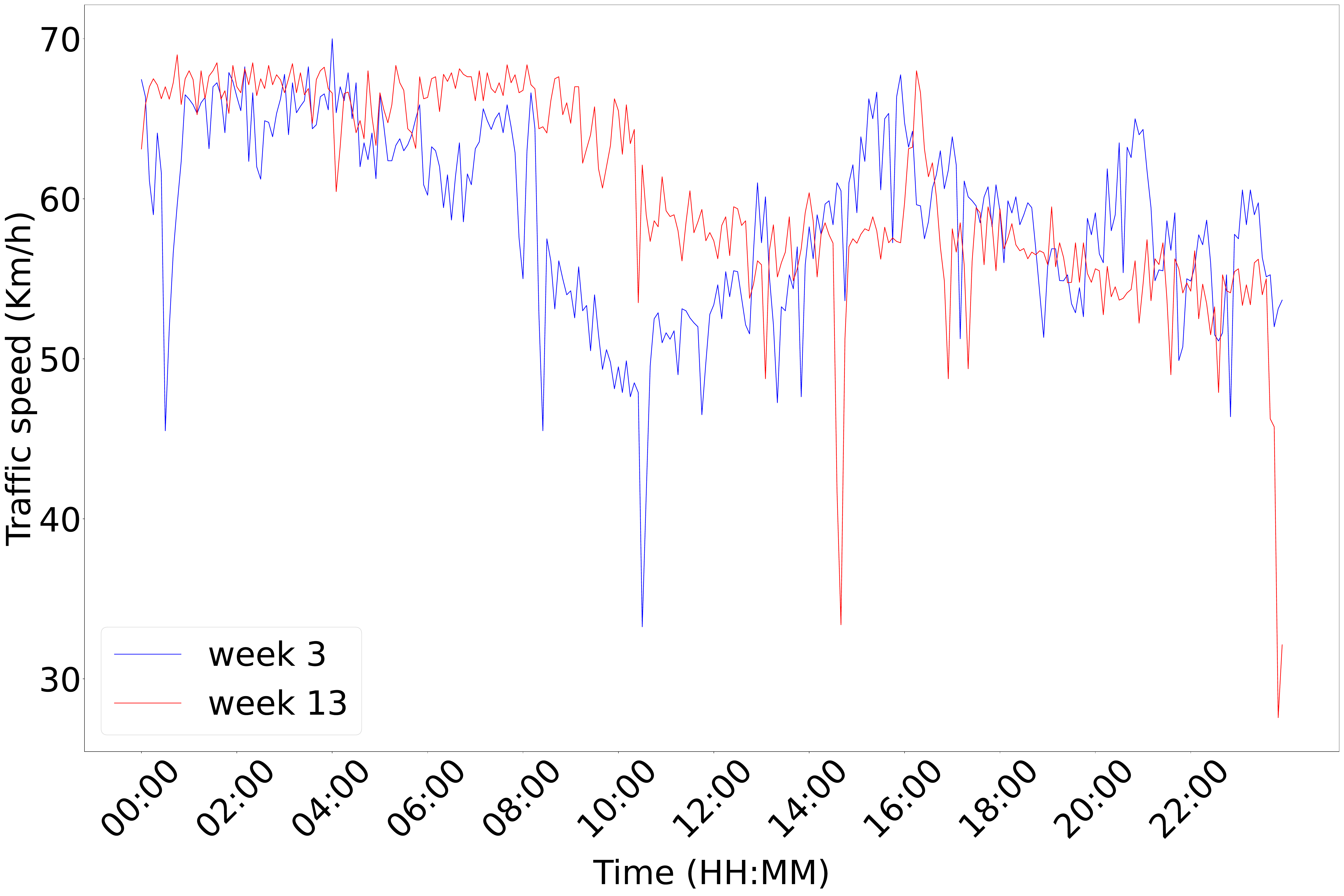}
        \caption{Traffic speed at node B.}
        \label{fig:OOD_2}
    \end{subfigure}
    \begin{subfigure}{0.3\linewidth}
        \centering
        \includegraphics[width=\textwidth]{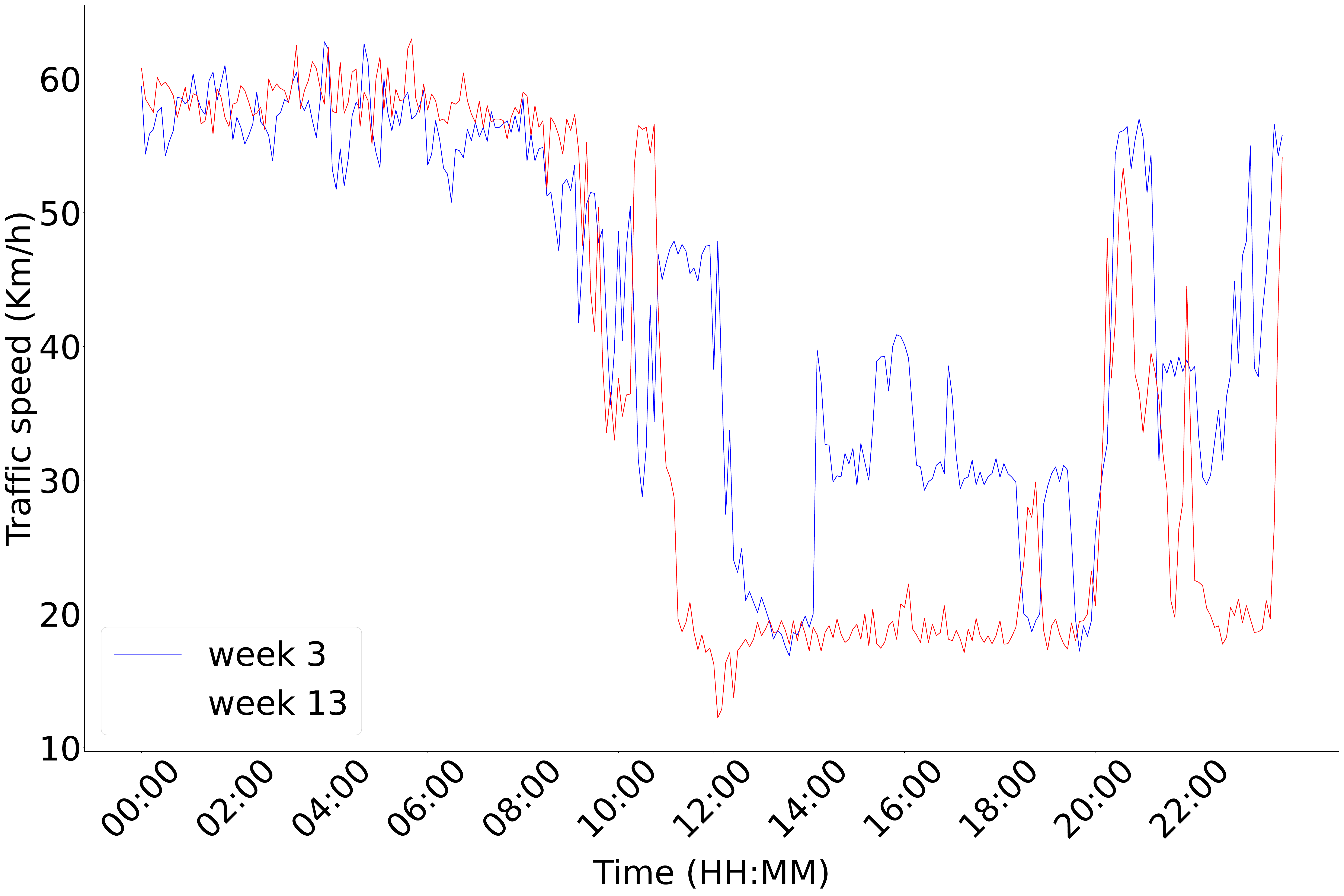}
        \caption{Traffic speed at node C.}
        \label{fig:OOD_3}
    \end{subfigure}
    \caption{The sampled traffic speed recorded by three sensors in the METR-LA dataset. The records correspond to two Wednesdays in Weeks 3 and 13 of the dataset.}
    \label{fig:OOD}
    \vspace{-5mm}
\centering
\end{figure*}

Urban flow data is typically structured as a spatial-temporal graph, whose nodes are commonly traffic sensors or geographical locations (e.g., grids \cite{wang2021gallat}), and are connected by edges based on physical proximity. 
Based on the time-dependent node features, the task is to predict the traffic flow associated with each node in the spatial-temporal graph at certain time steps. 
Given the characteristics of urban flow data, Spatial-Temporal Graph Neural Networks (STGNNs)~\cite{li2017diffusion,wu2019graph,yu2017spatio} have emerged as a natural choice to capture urban flow dynamics and make predictions based on historical observations. 
These models typically consist of two components: Graph Neural Networks (GNNs)~\cite{li2017diffusion,wu2019graph} for extracting spatial correlations between locations from the graph topology; and sequential models for capturing temporal evolutions, where typical examples include Recurrent Neural Networks (RNNs)~\cite{li2017diffusion,liu2023attention} and Temporal Convolution Networks (TCNs)~\cite{wu2019graph,yu2017spatio}. 
Along this line of work, some STGNNs further improve their predictive capability by introducing dynamic graph structures based on the similarity between temporal features of nodes ~\cite{bai2020adaptive,wu2020connecting,shang2021discrete},  or mining the complex combinatorial effect between spatial and temporal features ~\cite{ji2023spatio,cini2024taming}.

Despite the diversity in model designs, these models often assume that the urban flow data adheres to the independent and identically distributed (I.I.D.) assumption, which is challenging to guarantee in real-world applications. 
After deploying a model trained with historical observations, inference is usually performed with newly arrived, unseen data points, which may contain different patterns from the training data. 
This phenomenon is known as distribution shift, or out-of-distribution (OOD) during test time. 
In urban flow prediction, the spatial-temporal regularity can be easily disturbed by unexpected events like traffic accidents or extreme weather. Furthermore, during inference, it is impractical to assume any prior knowledge on the occurrence of such disturbing factors that lead to OOD data, weakening the accuracy of predictions and the utility of existing STGNNs.

Predicting urban flow with OOD data incurs non-trivial challenges. In Fig.~\ref{fig:OOD}, we provide a real example drawn from the same day (Wednesday) of two weeks from the METR-LA traffic dataset (see Section \ref{exp} for details), where node A (Fig.~\ref{fig:OOD_1}) is a traffic speed sensor, and nodes B (Fig.~\ref{fig:OOD_2}) and C (Fig.~\ref{fig:OOD_3}) are its two closest sensors. In the spatial-temporal graph, both B and C are one-hop neighbor nodes of sensor A. One direct observation is that, \textit{distribution shifts constantly happen within consecutive time intervals}, evidenced by the continuous traffic speed fluctuation over time. Although some STGNNs are able to capture periodical patterns as long-range dependencies (e.g., the same day of different weeks should have similar trends), \textit{distribution shifts that break such long-term regularity are still inevitable in complex urban flow data}. By comparing records from two Wednesdays, both nodes A and B are more congestive (i.e., lower speeds) between 8:00 and 12:00 in Week 3 than in Week 13, while the traffic speed between 12:00 and 20:00 in Week 13 is consistently lower than that in Week 3 at node C. As a result, in scenarios where Week 3 is a part of training data but the prediction is made with newly observed data in Week 13, the generalizability of STGNNs will be substantially limited.  

Furthermore, the \textit{heterogeneity of distribution shifts at different locations further creates an obstacle for learning useful spatial correlations} in the graph-structured urban flow data. As shown in Fig.~\ref{fig:OOD}, being the direct neighbors of node A, nodes B, and C exhibit distinct patterns when the new data drifts away from the earlier distribution. Since GNNs typically learn node representations by aggregating neighboring features in a predefined manner\cite{kipf2016semi,defferrard2016convolutional}, the impact from distribution shifts in nodes B and C will spill over onto node A, introducing more noisy signals for predicting its future traffic conditions. A straightforward remedy is to perform retraining every time new data points arrive to keep the STGNN model as up-to-date as possible. Unfortunately, this not only increases the computational costs linearly, but also challenges turnaround time in high-throughput applications (e.g., many traffic forecasting tasks \cite{li2017diffusion,shao2022spatial,liu2023spatio} only have a 5-minute interval). Thus, before entering the update cycle, the ideal STGNN should be capable of serving accurate predictions for a reasonable period of time by generalizing to changed data distributions. 

With the presence of distribution shifts in urban flow prediction, a key to enhancing the generalizability of STGNNs is to discover and leverage the invariant (i.e., causal) patterns within spatial-temporal data. 
Many studies on OOD generalization~\cite{rojas2018invariant,ahuja2020invariant} point out that distribution shifts are driven by the dynamics of underlying environments, where invariant risk minimization~\cite{arjovsky2019invariant,wu2022handling,liu2023flood} can be leveraged to optimize the model with augmented data drawn from diverse environments. However, directly augmenting urban flow data without substantial knowledge/heuristics on relevant external factors is prone to generating artificial distribution changes that are noisy and even misleading. 
As such, an alternative is to decouple invariant patterns from variant ones learned from the data~\cite{neuberg2003causality,pearl2016causal}. For example, when handling graph-structured data, \cite{zhang2022dynamic,chen2023causality,wu2022discovering} learn two disentangled graph structures that contain either invariant or variant connections between nodes. Unfortunately, these models are misaligned with urban flow prediction tasks as they only focus on a static graph topology that does not assume the temporal evolution of node features. 

To this end, we aim to build an OOD-robust STGNN that can distinguish invariant spatial-temporal patterns from urban flow data. 
Specifically, we propose Memory-enhanced Invariant Prompt learning (MIP), a novel solution to urban flow prediction.
In MIP, we attach a memory bank to the STGNN architecture, which learns and memorizes the causal patterns from the dynamic node features. Based on the information stored in the memory bank, a new graph structure reflecting the semantic causality between different locations is built, providing a complementary graph view to the default, distance-based graph structure for node representation learning. 
Then, the prompt vectors carrying invariant or variant patterns are extracted respectively by attentively querying the memory bank with each node's features. 
Furthermore, to facilitate end-to-end optimization via invariant risk minimization and ensure disentanglement between invariant and variant patterns, we put forward an innovative intervention pipeline that directly operates on the extracted variant prompts. Different from existing invariant learning methods  \cite{xia2024deciphering,zhou2023maintaining}, MIP bypasses the need for learning additional representations of different environments, and the designed intervention is a simple-yet-effective approach for implicitly mimicking the effect from data distribution shifts to node representations. The disentangled invariant patterns, along with both the geographical and semantic graphs, are eventually fed into a spatial-temporal backbone model to make accurate urban flow predictions.
To be concise, our contributions are summarized below:
\begin{itemize}
    \item \textbf{New Challenge.} We highlight a largely overlooked challenge in urban flow prediction: the pervasive presence of OOD data that hinders model generalizability. To address this, we propose a new framework, namely MIP to mitigate distribution shifts in urban flow prediction. 
    \item \textbf{New Method.} We extract invariant and variant features from a trainable memory bank and generate a supplementary graph structure. By implementing interventions on variant patterns and leveraging an invariant learning scheme, the invariant patterns are disentangled from the noisy data to facilitate accurate predictions.
    \item \textbf{SOTA Performance.} Extensive experiments on two real-world benchmark datasets have demonstrated the superiority of our method over state-of-the-art baselines when faced with OOD urban flow data.
\end{itemize}

\section{Related Work}
In this section, we review and summarize the research backgrounds that are relevant to our work.
\subsection{Deep Learning for Urban Flow Prediction} 
As a typical spatial-temporal prediction task, the most important part of urban flow prediction is to jointly learn time series features and topology features. Some early methods~\cite{schreiber2019long,yu2017deep,laptev2017time} utilize recurrent neural network (RNN) modules to learn temporal features from urban flow data. 
RNN-based models are adept at capturing temporal dependencies in sequential data; early methods, such as those by ~\cite{schreiber2019long,yu2017deep,laptev2017time}, utilize RNN modules to learn temporal features from urban flow data. RNN-based approaches often face challenges with long sequences, including inefficiencies and potential gradient issues when combined with graph convolution networks. Some models~\cite{zhang2016dnn,zhang2017deep} convert the road network to a two-dimensional grid and apply
traditional convolutional neural networks to predict urban flows.

Recently, spatial-temporal neural networks (STGNNs) have established themselves as state-of-the-art choices for urban flow prediction. STGNNs consist of GNN-based modules and sequential models that are alternately stacked, where typical variants include DCRNN~\cite{li2017diffusion}, GWNet~\cite{wu2019graph}, STGCN~\cite{yu2017spatio} and ST-MGCN~\cite{geng2019spatiotemporal}. Furthermore, attention mechanisms, including multi-head attention, are additionally used in fusing spatial and temporal information, such as GMAN~\cite{zheng2020gman}, ASTGCN~\cite{guo2019attention}, and PDFormer~\cite{pdformer}. Moreover, introducing some trainable features can also improve the performance of STGNNs, even with naive backbone models, such as STID~\cite{shao2022spatial}, STAEformer~\cite{liu2023spatio}, and MegaCRN~\cite{jiang2023spatio}. Besides, some physical theories can also guide spatial-temporal prediction, such as PGML~\cite{jiang2024physics} and STDEN~\cite{ji2022stden}.
However, these methods are designed based on the I.I.D assumption, making the extracted patterns solely dependent on the observed samples. Thus, these methods are prone to incorrect predictions when facing unobserved data with distribution shifts.

\subsection{Handling Out-of-Distribution (OOD) Data in Prediction} 
\label{OOD_related_work}
In the context of time series prediction where there are no spatial dependencies among observations, there are some representative studies that focus on OOD generalization. For example, 
CoST~\cite{woo2022cost} leverages inductive biases in the model architecture to learn disentangled seasonal and trend representations.
CaseQ~\cite{yang2022towards} is a novel hierarchical branching structure for learning context-specific representations of sequences. It could dynamically evolve its architecture to adapt to variable contexts. 
AdaRNN~\cite{du2021adarnn} splits historical time sequences into different classes with large distribution gaps and dynamically matches input data to these classes to identify contextual information. Dish-TS~\cite{fan2023dish} separately learns the distribution of input and output spaces, which naturally captures the distribution difference. However, these models do not account for spatial information and thus are incompatible with urban flow prediction. 

To tackle spatial distribution shifts, graph-based invariant learning is a common practice. It focuses on extracting features that are consistent across various conditions, which are then incorporated into the prediction model to enhance its accuracy. According to causal theory~\cite{pearl2016causal,neuberg2003causality}, previous studies~\cite{zhang2022dynamic,wu2022discovering} assume data always contain both invariant and variant patterns, while graph models sometimes make predictions based on the variant part and would fail on OOD data. Some models learn a mask over the graph adjacency matrix to separate the invariant and variant parts, such as DIDA~\cite{zhang2022dynamic}, DIR~\cite{wu2022discovering}, and CIE~\cite{chen2023causality}, where the disentangled invariant patterns are fed into an auxiliary model to facilitate accurate predictions. By performing interventions with the extracted variant patterns, the model is trained to better distinguish the invariant predictive signals from variant ones. Despite their applicability to graph-structured data, these models only consider static graphs without temporal evolutions, highlighting the need for designing invariant learning methods for spatial-temporal graphs. 

There are some models dedicated to overcoming distribution shifts in spatial-temporal data. 
For example, CaST~\cite{xia2024deciphering} disentangles the environmental feature and the entity feature based on causal treatments~\cite{neuberg2003causality}, and it replaces the environment feature with the vector closest to it in the environment codebook, which contains $K$ vectors and each of them represents an environment. CauSTG~\cite{zhou2023maintaining} designs a hierarchical invariance explorer, which merges the models trained across various environments. Concretely, it presets there are $K$ virtual environments, and $M$ models are trained within each environment. Firstly, in each environment, the $M$ models are locally merged as one, and it obtains $K$ models for $K$ environments, and then it globally merges them to obtain the final prediction model.
STONE~\cite{wang2024stone} learns both spatial and temporal similarity matrices as adjacency matrices for STGNN to make predictions. It then generates $K$ masks to selectively remove connections between spatial nodes in the graph, so as to simulate the effect of different environments.
However, these aforementioned methods require learning and generating environments to facilitate accurate predictions, which heavily rely on specifically designed model mechanisms and are highly sensitive to the number of virtual environments $K$. If more different environments are set, they are likely to introduce noise to the model parameters, while a smaller $K$ will make the data sampled from different environments indistinguishable. As the urban flow data evolves over time, the underlying environments are also intrinsically dynamic, which are unable to be captured via a fixed number of learned environments.

\section{Preliminaries}
We hereby introduce some key preliminaries in the context of this work. 
\label{preliminary}
\subsection{Problem Formulation}
In urban flow data, a geolocation graph can be defined as: $\mathcal{G}=\{\mathcal{V}, \mathcal{E}\}$, where $\mathcal{V}$ is the set of $N$ nodes and $\mathcal{E}$ is the set of edges. Usually, the nodes represent sensors or geographical regions that carry continuous traffic records, and an edge will connect two nodes if their physical distance is smaller than a predefined threshold. Moreover, $\mathbf{A} \in \{0,1\}^{N \times N}$ is the adjacency matrix derived from the graph, where for node indexes $n,v\leq N$, each entry $\mathbf{A}[n,v]=1$ if $(n, v) \in \mathcal{E}$
and $\mathbf{A}[n,v]=0$ if $(n, v) \notin \mathcal{E}$. At each time step $t\leq T$, all nodes' dynamic features are represented via a matrix $\mathbf{X}^t \in \mathbb{R}^{N \times k}$, with $k$ representing the dimensionality of time-varying features. In practice, $\mathbf{X}^t[n]\in \mathbb{R}^{k}$ encodes $k$ observed urban flow signals (e.g., in- and out-flows) of node $n$ at time $t$. Note that for urban flow prediction tasks, only the node features change over time, whereas the geolocation graph structure stays unchanged \cite{li2017diffusion,yu2017spatio,liu2022msdr,liu2023attention}. Following the commonly adopted setting \cite{wu2019graph}, given the observed $T$ historical observations $\{\mathbf{X}^{t}\}_{t=1}^T$ and the geolocation graph $\mathcal{G}$, the task objective is to train a model that predicts the next $T$ urban flow signals $\{\mathbf{X}^{t'}\}_{t'=T+1}^{2T}$:
\begin{equation}
\mathbf{Y} \simeq f_{\theta}(\mathbf{X}, \mathcal{G}),
\end{equation}
where $\mathbf{X}, \mathbf{Y}\in \mathbb{R}^{T\times N\times k}$  are respectively the tensorized versions of input $\{\mathbf{X}^{t}\}_{t=1}^T$ and output $\{\mathbf{X}^{t'}\}_{t'=T+1}^{2T}$, and $f_{\theta}(\cdot)$ is the prediction model parameterized by $\theta$. 

Usually, the optimization of $\theta$ is based on the I.I.D assumption, which means the training data and the testing data are drawn from the same distribution. In practice, this assumption can hardly be guaranteed as the training and testing data are drawn from different environments $E$. 
In this study, we aim to develop a model with high generalization capabilities when the test data is OOD. Specifically, the optimal model parameter $\theta^*$ should achieve minimal risk on an OOD test set:
\begin{equation}
\begin{aligned}
&\min\mathbb{E}_{(\mathbf{X}', \mathbf{Y}')\sim p(\mathbf{X}',\mathbf{Y}'|E_{test})} \mathcal{L}(f_{\theta}(\mathbf{X}',\mathcal{G}),\mathbf{Y}'), \\
s.t.\,\, \theta^*=& \text{arg}\min_\theta\mathbb{E}_{(\mathbf{X}, \mathbf{Y})\sim p(\mathbf{X},\mathbf{Y}|E_{train})} \mathcal{L}(f_{\theta}(\mathbf{X},\mathcal{G}),\mathbf{Y}),\\
\end{aligned}
\label{eq:OOD_object}
\end{equation}
where $E_{train} \ne E_{test}$ and $\mathcal{L}(\cdot)$ quantifies the prediction error.

\subsection{Invariant Learning under Distribution Shifts}
\label{ilearning}

As the traffic flows between geographically connected nodes, it necessitates the use of a graph neural network (GNN) to explore the spatial-temporal interactions among different nodes. Intuitively, GNNs learn a node's representation by propagating features from locally connected nodes. Denoting an arbitrary GNN model as $m$, for a specific node $n$, its representation learning process at time $t$ can be abstracted as the following:
\begin{equation}
    F^t(n) = m(\mathcal{G}_n, \mathbf{X}^t_{\mathcal{G}_n}),
\end{equation}
where $F^t(n)$ denotes the time-dependent latent features of node $n$, $\mathcal{G}_n\in \mathcal{G}$ is $n$'s ego graph that consists of node $n$ and its neighbour nodes within predefined hops, and $\mathbf{X}^t_{\mathcal{G}_n}$ denotes the node features correspond to $\mathcal{G}_n$ at time $t$.
The model $m$ is designed to capture both spatial and temporal features from the ego graph. Drawing on causality theory~\cite{pearl2016causal,neuberg2003causality}, several studies~\cite{zhang2022dynamic,wu2022discovering} propose the following assumption:

\textbf{Assumption 1:} \textit{For an arbitrary node $n$ and time $t$, given $\mathcal{G}_n, \mathbf{X}^t_{\mathcal{G}_n}$ and the corresponding label $y^t_n$ drawn from any distribution, the extracted features $F^t(n)$ contains both variant patterns $F^t_V(n)$  and invariant patterns $F^t_I(n)$. Then, there exists a prediction function $pred(\cdot)$, for which the invariant feature $F^t_I(n)$ is sufficiently predictive for label $y^t_n$ and the variant feature $F^t_V(n)$ does not hold causation to $y^t_n$, i.e., i.e., $y^t_n\simeq pred(F^t_I(n),\mathcal{G}_n)$. By using $|$ and $\perp$ to respectively denote dependent and non-dependent relationships, this assumption can be expressed as ${y}^t_n \perp {F}^t_V(n) | {F}^t_I(n)$}. 

\begin{figure*}[t!]
    \centering
    \includegraphics[width=0.92\textwidth]{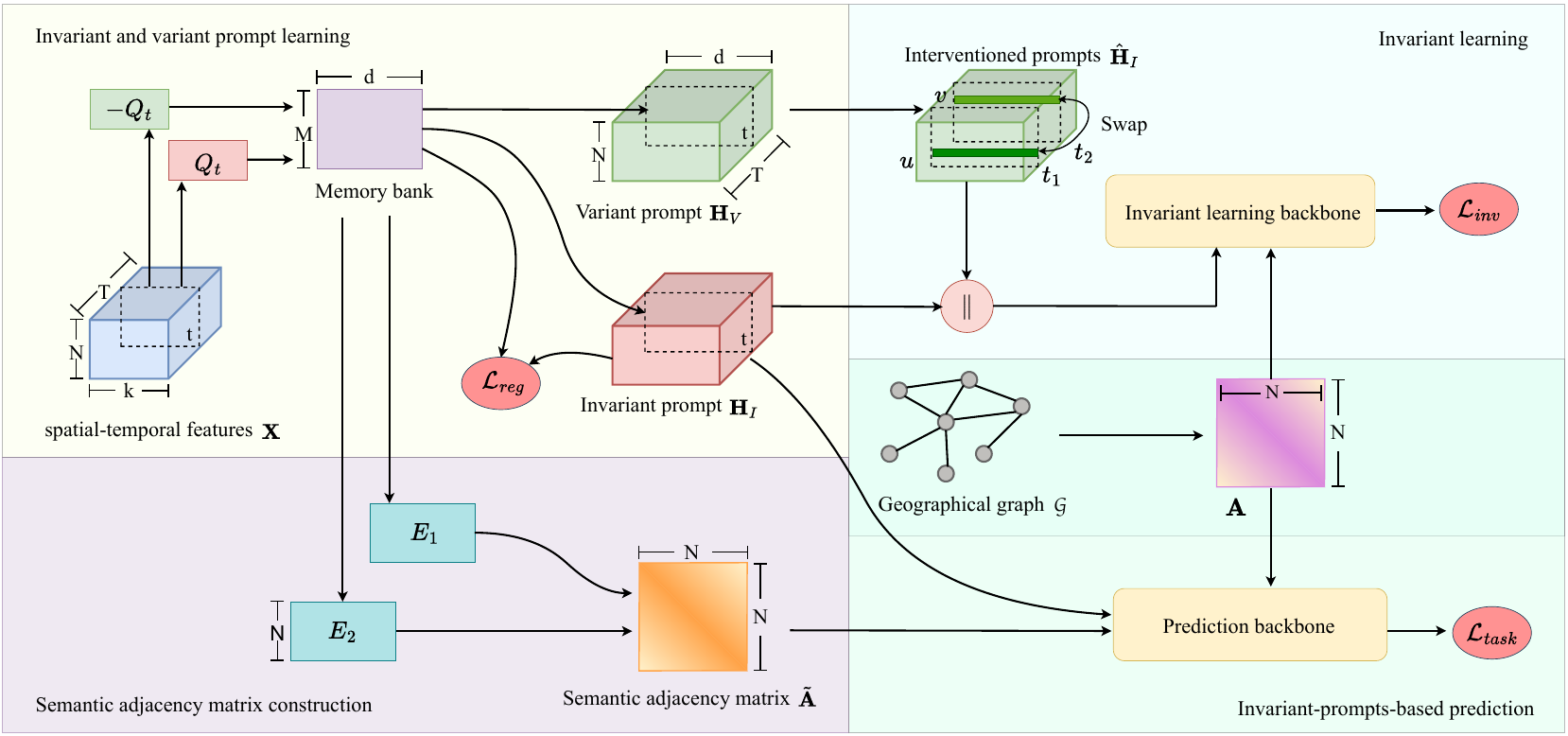}
    \caption{\small The workflow of MIP.}
    \label{fig:frame}
    \vspace{-3mm}
\end{figure*}

This assumption is represented through a Structural Causal Model (SCM) depicted in Fig.\ref{fig:scm}, where each arrow in the SCM illustrates a relationship between two variables. Typically, $F^t(n)$ extracted from $\mathcal{G}^t_v$ contains both variant and invariant features (line 0), jointly driving the prediction $pred(F^t(n))$ to approximate $y^t_n$ (line 2). By disentangling the effect of invariant feature $F^t_I(n)$ (line 1) with variant feature $F^t_V(n)$ (line 3), we aim to remove the spurious relationship between $F^t_V(n)$ and label $y^t_n$ (line 3), and only build the predictor based on the causal, invariant feature $F^t_I(n)$ (line 1). 
Based on this analysis, we rewrite our graph-level objective below:
\begin{equation}
\begin{aligned}
&\text{arg}\min_{\theta_1,\theta_1}\mathbb{E}_{(\mathbf{X},\mathbf{Y})\sim p(\mathbf{X},\mathbf{Y} |E_{train})} \mathcal{L}(f_{\theta_1}(\mathbf{H}_I,\mathcal{G}), \mathbf{Y}), \\
s.t. \,\, & \mathbf{H}_I,\mathbf{H}_V=\{{F}^t_I(n)\}_{t,n=1}^{T,N},\{{F}^t_V(n)\}_{t,n=1}^{T,N}=f_{\theta_2}(\mathbf{X}), \\ 
& \mathbf{Y}\perp \mathbf{H}_V | \mathbf{H}_I,
\end{aligned}
\end{equation}
where we use $\mathbf{H}_I, \mathbf{H}_V\in \mathbb{R}^{T\times N \times d}$ to respectively denote all $T\times N$ variant and invariant patterns extracted from input $\mathbf{X}$, and $d$ is the latent dimensionality of both extracted patterns. Here, $f_{\theta_2}(\cdot)$ is the invariant learning backbone model that disentangles invariant patterns with the variant ones from the dynamic node features, and $f_{\theta_1}$ is the prediction backbone model that is only fed with the invariant patterns to derive final predictions. 
Based on the formulation, a key step is to train $\psi_{\theta_2}$ towards distinguishing invariant and variant patterns. To achieve this, based on the interventional distribution in causality theory~\cite{wu2022discovering}, the objective can be converted into the following invariant learning loss:
\begin{equation}
\begin{aligned}\min_{\theta_2}\,& \mathbb{E}_{(\mathbf{X}_s,\mathbf{Y})\sim p(\mathbf{X}_s,\mathbf{Y} |E_{s})}\mathcal{L}(pred(f_{\theta_2}(\mathbf{X}_s),\mathcal{G}), \mathbf{Y}) \\
    +&\lambda Var_{(\mathbf{X}_s,\mathbf{Y})\sim p(\mathbf{X}_s,\mathbf{Y} |E_{s})}\mathcal{L}(pred(f_{\theta_2}(\mathbf{X}_s),\mathcal{G}), \mathbf{Y}),
    \label{eq:dirloss}
\end{aligned}
\end{equation}
where $E_s\neq E_{train}$ denotes an intervention on the original environment that leads to features $\mathbf{X}_s\in \mathbb{R}^{T\times N \times k}$ with shifted distributions, and $s\in \mathcal{S}$ is sampled from an intervention set. $pred(\cdot)$ denotes a predictor that uses both invariant and variant patterns emitted by $f_{\theta_2}(\mathbf{X}_s)$ to predict the ground truth label, which does not necessarily share the same structure or parameterization with $f_{\theta_1}(\cdot)$. Note that $E_s$ only performs intervention on variant patterns, leaving the invariant features unaffected. The first term minimizes the prediction loss, whereas in the second term, $Var$ denotes the variance and $\lambda$ is a balancing hyperparameter. As the variant patterns are unrelated to the label $\mathbf{Y}$, the prediction should remain stable regardless of the intervened variant patterns introduced by $E_s$, leading to a lower variance. Furthermore, as the variance needs to be kept small during training, this term also helps prevent the intervention from deviating too far from realistic distributions. Through iterative environmental interventions, $\psi_{\theta_2}$ is trained to differentiate invariant patterns $\mathbf{H}_I$ and variant patterns $\mathbf{H}_V$, thus reinforcing the causal relationship between $\mathbf{H}_I$ and the urban flow ground truth $\mathbf{Y}$.  

\section{MIP: The Proposed Method}
In this section, we introduce a universal framework named Memory-enhanced
Invariant Prompt learning (MIP) for urban flow prediction under OOD scenarios, whose main components are depicted in Figure 1. In what follows, we unfold the design of MIP by introducing the design of the memory bank, as well as the backbones for invariant learning (i.e., $f_{\theta_2}(\cdot)$) and spatial-temporal prediction (i.e., $f_{\theta_1}(\cdot)$). 

\subsection{Memory Bank and Semantic Graph Construction}
\label{nodebank}
A key advantage of our MIP is that, instead of generating intervened environment $E_s$ and simulating subsequent changes in the learned latent patterns, MIP directly intervenes the latent space to mimic the changes in the learned representations after environmental intervention. To facilitate this, we first need to mine the latent patterns correlated with the predicted labels from the time-varying node features. 
Therefore, we extract and store these representative causal features with a memory bank~\cite{jiang2023spatio}. 
The memory bank can be represented as $\mathbf{\Phi} \in \mathbb{R}^{M \times d}$, where $M$ and $d$ represent the number of virtual nodes and their dimensions, respectively. Essentially, each of the $M$ virtual nodes in the memory bank is assigned a $d$-dimensional prototype vector $\mathbf{\Phi}[m]\in \mathbb{R}^d$ ($m\leq M$) that summarizes a part of the latent, invariant features within the spatial-temporal node features $\mathbf{X}$. The memory bank $\mathbf{\Phi}$ supports two subsequent computations: generating variant and invariant prompts through a querying process, which is covered in Section \ref{learn_prompts}; and providing a semantic graph in addition to the geographical graph for learning a node's spatial-temporal representation, which we will introduce below. 

In urban flow prediction tasks, constructing the original graph solely based on physical distances between nodes can introduce biases during GNN's information propagation. This is because geographic proximity does not necessarily imply similar temporal patterns, especially in OOD scenarios. For instance, consider two connected nodes representing adjacent city blocks. If one block experiences a traffic accident leading to road closure and reduced traffic flow, it does not mean the neighboring block will exhibit a similar pattern. Instead, the adjacent block is likely to experience increased traffic flow as people will seek alternative routes. To address this limitation, we introduce an auxiliary graph based on the semantic distance between causal node representations, which are constructed from highly invariant features from the memory bank $\mathbf{\Phi}$. This semantic graph is able to complement the geolocation-based graph, 
thus providing additional predictive signals.
This memory bank-based semantic graph is constructed as the following:
\begin{align}
\begin{aligned}
&\mathbf{E}_1 = \mathbf{W}_A   \mathbf{\Phi}, \\
&\mathbf{E}_2 = \mathbf{W}_B   \mathbf{\Phi}, \\
&\tilde{\mathbf{A}} = \text{softmax} \left( \mathbf{E}_1   \mathbf{E}^{\top}_2 \right), \label{eq:adpA}
\end{aligned}
\end{align}
where $\mathbf{W}_A,\mathbf{W}_B \in \mathbb{R}^{N \times M}$ are trainable projection matrices that map the $M$ prototype vectors in the memory bank into $N$ node representations $\mathbf{E}_1$ and $\mathbf{E}_2$. The row-wise softmax is responsible for scaling each node's similarity with other nodes into a distribution. As the memory bank already encapsulates critical information of the urban flow, the newly developed semantic adjacency matrix provides additional information propagation channels between nodes. Notably, different from variants that build a semantic graph for each time step based on extracted node features, our semantic graph is solely based on the memory bank. As the memory bank stays fixed and is trained to contain only invariant patterns, the same semantic graph is shared across all time steps for node representation learning. This design not only ensures efficiency, but also helps with our model's robustness in OOD scenarios as the causal edges in the memory bank-based semantic graph are free from fluctuations caused by the time-varying node features. 

\subsection{Learning Invariant and Variant Prompts}
\label{learn_prompts}
As a core part of OOD generalization, we discuss given all nodes' temporal features $\mathbf{X}^t\in \mathbb{R}^{N\times k}$ correspond to time $t$, how to separate causal and spurious patterns -- which we term invariant and variant prompts in this work. 
Firstly, we project the input $\mathbf{X}^t$ into a query matrix $\mathbf{Q}_t\in \mathbb{R}^{N\times d}$:
\begin{equation}
\mathbf{Q}_t = \mathbf{X}_t   \mathbf{W}_Q + \mathbf{b}_Q ,
\label{eq:Qt}
\end{equation}
where $\mathbf{W}_Q \in \mathbb{R}^{k \times d}$ and $\mathbf{b}_Q \in \mathbb{R}^{d}$ are trainable parameters of the linear layer. As $\mathbf{Q}_t$ contains both invariant and variant patterns, we further disentangle them by querying the invariant memory bank.
To obtain invariant prompts, we multiply the query matrix $\mathbf{Q}_t$ with the memory bank $\mathbf{\Phi}$ to obtain a prompt score $\mathbf{S}^t_I\in \mathbb{R}^{N \times M}$, and then get the invariant prompt by aggregating the memory bank items with this score. This process can be formulated as:
\begin{equation}
\begin{aligned}
&\mathbf{S}^t_I = \text{softmax} \left(\mathbf{Q}_t \mathbf{\Phi}^{\top} \right), \\
&\mathbf{H}^t_I = \mathbf{S}^t_I \mathbf{\Phi}, \label{eq:Hinv}
\end{aligned}
\end{equation}
where $\mathbf{H}^t_I\in \mathbb{R}^{N\times d}$ is the computed invariant prompt. Similarly, the variant prompt can be extracted in an analogous process, with a minor modification: 
\begin{equation}
\begin{aligned}
&\mathbf{S}^t_V = \text{softmax} \left( -1   \mathbf{Q}_t \mathbf{\Phi}^{\top} \right), \\
&\mathbf{H}^t_V = \mathbf{S}^t_V \mathbf{\Phi},\label{eq:Hvar2}
\end{aligned}
\end{equation}
where a negation is applied before the softmax function when calculating $\mathbf{S}^t_V$, so as to flip the score distribution and assign higher weights to invariant patterns that are less relevant to the memory $\mathbf{\Phi}$. 
As this is executed for all time steps, we can obtain a sequence of $T$ prompts $\{\mathbf{H}^{1}_I, \mathbf{H}^{2}_I,...,\mathbf{H}^{T}_I\}$ and $\{\mathbf{H}^{1}_V,\mathbf{H}^{2}_V,...,\mathbf{H}^{T}_V\}$. By respectively concatenating invariant and variant prompts across time, we can obtain two prompt tensors $\mathbf{H}_I, \mathbf{H}_V \in \mathbf{R}^{T\times N \times d}$ for subsequent computations. 

\subsection{Intervention Mechanism}
\label{intervention}
Given the analysis in Section~\ref{ilearning}, the data $\mathbf{X}$ comprises invariant and variant components that are respectively encoded into $\mathbf{H}_I$ and $\mathbf{H}_V$, where $\mathbf{H}_V$ is associated with the environment but unrelated to the label $Y$. As per our discussions earlier, the key to OOD generalization is to ensure that the invariant, causal patterns $\mathbf{H}_I$ are fully distinguished from the variant ones $\mathbf{H}_V$. To facilitate this, a common practice is to alter the input data's distribution with interventions on the environment, from which multiple versions of $\mathbf{H}_V$ can be extracted. Intuitively, through invariant risk minimization, the model is able to deliver consistent predictions with minimal variance regardless of the environments. That is, a sufficiently informative $\mathbf{H}_I$ can be isolated from the noises in $\mathbf{H}_V$, such that the predictions are always based on the causal patterns in $\mathbf{H}_I$. However, directly intervening in the environments is a less favorable option for urban flow prediction tasks, as this intervention mechanism \cite{wu2022discovering,xia2024deciphering,wu2024graph} commonly needs to additionally parameterize and learn the underlying environments in order to alter the raw data distribution. Consequently, this brings a dilemma where a fixed number of environments across all time steps can hardly capture the evolving nature of such spatial-temporal graphs, while simulating a set of environments for every time step can in turn introduce noises and challenge efficiency.

As such, we innovatively propose to generate spatial-temporal interventions in the latent space, which more effectively mimics the changes in the learnable patterns after possible distribution shifts within the input data. 
Specifically, given the invariant and variant prompts $\mathbf{H}_I, \mathbf{H}_V$ extracted from $\mathbf{X}$, 
we exchange a predefined ratio of features within $\mathbf{H}_V$ between different nodes and time points. The details of spatial-temporal intervention are described in Algorithm 1. With Algorithm 1, we simplify the intervention process into the following:
\begin{equation}
    \hat{\mathbf{H}}_V = \textsc{Intervene}(\mathbf{H}_V, r),
\end{equation}
where $\hat{\mathbf{H}}_V$ denotes the intervened variant prompts after feature exchanges have taken place for $rN$ nodes in the graph. Note that, the swap is not constrained to node features at the same time step, so as to account for the spatial-temporal fluctuations within the variant patterns. Also, by producing intervened variant prompts with representations learned from the original input data, the generated $\hat{\mathbf{H}}_V$ remains plausible and challenging for refining the invariant prompts $\hat{\mathbf{H}}_I$.

\begin{algorithm}[t!]
\caption{$\textsc{Intervene}(\mathbf{H}_V, r)$}
\begin{algorithmic}[1] 
\STATE \textbf{Input}: Variant prompt tensor $\mathbf{H}_V\in \mathbb{R}^{ T\times N\times d}$, ratio $r$
\STATE \textbf{Output}: Intervened variant prompt $\hat{\mathbf{H}}_V\in \mathbb{R}^{ T\times N\times d}$
\STATE $\hat{\mathbf{H}}_V\leftarrow \mathbf{H}_V$
\FOR{s=1,2,...,$\lfloor \frac{rN}{2} \rfloor$}
\STATE Randomly sample a node pair $(w,v)$ s.t. $w,v\in [1,N]$;
\STATE Randomly select a time step pair $(i,j)$ s.t. $i,j\in [1,T]$;
\STATE $\hat{\mathbf{H}}_V[i,w]\leftarrow\mathbf{H}_V[j,v]$, $\hat{\mathbf{H}}_V[j,v]\leftarrow\mathbf{H}_V[i,w]$;
\ENDFOR
\RETURN $\hat{\mathbf{H}}_V$
\label{alg:intervention}
\end{algorithmic}
\end{algorithm}

\subsection{Invariant Learning }
After obtaining the intervened variant prompts $\hat{\mathbf{H}}_V$, we are able to train the prompt extractor described in Section \ref{learn_prompts} via invariant learning, so as to distinguish the invariant and variant prompts. To achieve this, the invariant and the intervened variant prompts are concatenated and then input into a supplementary predictor $pred(\cdot)$ to generate predictions: \begin{equation}
    \tilde{\mathbf{Y}} = pred(\mathbf{H}_I||\hat{\mathbf{H}}_V,\mathbf{A}),
\end{equation}
where $||$ denotes tensor concatenation along the last dimension, $\tilde{\mathbf{Y}} \in \mathbb{R}^{T \times N\times k}$ is the predicted urban flow at all locations and time steps. The choice of $pred(\cdot)$ is flexible with most STGNNs. Notably, $pred(\cdot)$ is only responsible for differentiating invariant and variant prompts and will not be used for computing the final predictions. Hence, we adopt   GWNet~\cite{wu2019graph}, a simple yet effective STGNN to serve as $pred(\cdot)$. 


For training, we first define the loss for a single node $n$ at one time step $t$:
\begin{equation}
    l(t,n) = \frac{1}{k}\sum_{k'=1}^{k}|\tilde{\mathbf{Y}}[t,n,k']-\mathbf{Y}[t,n,k']|,
\end{equation}
based on which the invariant learning loss is defined:
\begin{equation}
\begin{aligned}
    \mathcal{L}_{inv} &=\mathbb{E}_{(t,n))}l(t,n)  +
    \lambda_1 Var_{(t,n)} l(t,n),\\
    &t\in [1,T] ,\,\,\,\,n \in [1,N],\\
\end{aligned}
\end{equation}
where the first and second terms respectively reduce the mean and variance of the prediction error across locations and time steps. More specifically, the first term ensures that $pred(\cdot)$ is optimized towards correctly predicting the urban flow in different environments with $\mathbf{H}_I$, while the second term enforces that when the predictions are conditioned on $\mathbf{H}_I$, there are minimal performance fluctuations despite the presence of noisy signals carried by the intervened variant prompts $\mathbf{H}_V$. 

\subsection{Backbone Model for Spatial-Temporal Prediction}
\label{backbone}
Once the invariant features $\mathbf{H}_I$ are extracted with the invariant learning process, a spatial-temporal backbone model needs to be in place for producing the final predictions. 
In MIP, our backbone model consists of alternately stacked GNN layers and temporal Transformer layers.

\noindent\textbf{GNN Layer.} 
The GNN layer is fed with both the geographical and semantic adjacency matrices $\mathbf{A}$, $\tilde{\mathbf{A}}$ and the invariant prompts $\mathbf{H}_I$ to learn node representations with information propagation. 
Since the geographical adjacency matrix $\mathbf{A}$ is symmetric and hardly captures the directed nature of interactions in urban flow data, we derive forward and backward transition matrices from $\mathbf{A}$ through a bidirectional, degree-weighted random walk process \cite{li2017diffusion}:
\begin{equation}\mathbf{P}_f=\mathbf{D}^{-1}\mathbf{A},\,\,\mathbf{P}_b=(\mathbf{D}^\top)^{-1}\mathbf{A}^\top,
\end{equation}
where $\mathbf{D}$ is the degree matrix of $\mathbf{A}$. 
By incorporating the semantic  adjacency matrix, the propagation process in the GNN from layer $l$ to $l+1$ is summarized as follows:
\begin{equation}
\mathbf{G}_{l+1}^t = \sum_{z=0}^{Z} ( \mathbf{P}^{z}_f\mathbf{G}_l^t\mathbf{W}^{z}_{1} +  \mathbf{P}^{z}_b\mathbf{G}_l^t\mathbf{W}^{z}_{2}+ \tilde{\mathbf{A}}^{z}\mathbf{G}^t_l\mathbf{W}^{z}_{3}),
\end{equation}
where $z\leq Z$ controls the order of the information propagation, and $\mathbf{W}^{(k)}_{\cdot} \in \mathbb{R}^{d\times d}$ denotes the learnable weights. Notably, the GNN layer only processes one graph snapshot at a time, and the initial node embeddings are set to $\mathbf{G}_0^t=\mathbf{H}_V^t$ when $l=0$. 


\begin{table}[]
\caption{Statistics of the datasets.}
\centering
\footnotesize
\resizebox{0.8\linewidth}{!}{
\begin{tabular}{llll}
\toprule
Dataset       & METR-LA     &  NYCBike1     \\
\midrule
Start Time    & 2012/3/1    &  2014/4/1     \\
End Time      & 2012/6/30   &  2014/9/30    \\
Time Interval & 5 minutes   &  1 hour       \\
Timesteps     & 34,272      &  4,392        \\
Spatial Units & 207 sensors &  8$\times$16 regions \\
\bottomrule
\end{tabular}
}
\vspace{-5mm}
\label{tab:data}
\end{table}

\begin{table*}[]
\caption{Performance comparison results. The best results are marked in bold and the second best results are underlined.}
\centering
\resizebox{0.85\linewidth}{!}{
\begin{tabular}{l|lll|lll|lll|lll}
\toprule
\multirow{2}{*}[-2pt]{METR-LA}    & \multicolumn{3}{c|}{testing set 0}                & \multicolumn{3}{c|}{testing set 1}                & \multicolumn{3}{c|}{testing set 2}  & \multicolumn{3}{c}{overall results}               \\
\cmidrule{2-4} \cmidrule{5-7} \cmidrule{8-10} \cmidrule{11-13}
    & MAE  & RMSE  & MAPE    & MAE   & RMSE  & MAPE    & MAE  & RMSE  & MAPE & MAE   & RMSE  & MAPE    \\
\midrule  
STGCN      & 3.33                   & 7.15  & 10.14\% & 3.63                   & 7.47  & 10.90\% & 3.63                   & 7.58  & 10.21\% &3.53  &  7.40   &  10.42\% \\
DCRNN      & 3.33                   & 7.28  & 10.01\% & 3.58                   & 7.49  & 10.80\% & 3.69                   & 7.87  & 10.41\%  &  3.53    &   7.55 &  10.41\%\\
STNorm     & 3.33                   & 7.17  & 10.09\% & 3.65                   & 7.57  & 11.16\% & 3.63  & 7.61  & 10.23\% &  3.53 & 7.45  & 10.49\%  \\
GMSDR      & \uline{3.27}           & \uline{6.99}  & 9.75\%  & \textbf{3.49}          & \uline{7.36}  & 10.83\% & \textbf{3.50}                   & \uline{7.47}  & 10.01\% & \textbf{3.42}  & \uline{7.27}  &  \uline{10.20\%} \\
MegaCRN    & \textbf{3.22}          & 7.05  & 9.69\%  & 3.64                   & 7.65  & 11.04\% & 3.79                   & 8.00  & 10.75\% &3.55  &  7.57  &  10.49\% \\
CauSTG     & 3.33                   &7.08  &  9.86\%& 3.64&7.44   & 10.81\% &                   3.66&   7.55& 10.10\%  & 3.55 &  7.36  &  10.26\%   \\
TESTAM    &3.36   &  7.33&  \uline{9.56}\%&  3.62 &  7.58  & \textbf{10.26\%}&  3.69 & 7.89  & \uline{9.98\%} & 3.56  &  7.60   &  9.93\% \\ 
MIP & 3.28                   & \textbf{6.87}  & \textbf{9.52\%}         & \uline{3.55}  & \textbf{7.19}  & \uline{10.40\%} & \uline{3.57}          & \textbf{7.28}  & \textbf{9.73\%}  &  \uline{3.46}   &  \textbf{7.11}  &  \textbf{9.88\%}  \\
\midrule  
\multirow{2}{*}[-2pt]{NYCBike(In)}   & \multicolumn{3}{c|}{testing set 0}                & \multicolumn{3}{c|}{testing set 1}                & \multicolumn{3}{c|}{testing set 2}    & \multicolumn{3}{c}{overall results}             \\
\cmidrule{2-4} \cmidrule{5-7} \cmidrule{8-10}\cmidrule{11-13}
    & MAE                    & RMSE  & MAPE    & MAE                    & RMSE  & MAPE    & MAE                    & RMSE  & MAPE   & MAE   & RMSE  & MAPE   \\
\midrule  
STGCN      & 4.90                  & 8.41 & 50.38\% &                   4.69&7.55 & 51.11\% &                  5.33 & 9.42 & 63.05\% &  4.97  & 8.46  & 54.84\%  \\
DCRNN      & 6.24                  & 10.04 & 80.58\% & 5.90                  & 9.18 & 77.03\% & 6.47                  & 10.74 & 92.59\%  & 6.20  &  9.99  & 83.40\% \\
STNorm     & 4.83          & 8.52 & \uline{46.49\%} & \textbf{4.53}         & 8.29 & 50.51\% & \uline{5.28}         & \uline{9.37} & \uline{56.21\%} &\uline{4.88} & \uline{8.38}  & \uline{49.35\%}  \\
GMSDR      & 5.10                  & 8.96 & 48.70\% & 4.86                  & 8.14 & \uline{49.14\%} & 5.41   & 9.68 & 61.02\%  & 5.12 & 8.92  &  52.95\%\\
MegaCRN    & \textbf{4.62}  & \textbf{7.96} & 46.65\% & 5.15                  & 8.87 & 55.35\% & 5.62                  & 9.51 & 65.81\% & 5.13  &  8.78  &  55.93\% \\
CauSTG      & 4.95         & 8.51 & 49.21\% & 4.83  & \uline{7.91} & 49.47\% & 5.37& 9.45 & 59.73\%  &  5.05  &  8.63  &  52.80\%\\
TESTAM     & 5.06         & 8.51 & 47.69\% & 5.18  & 8.38 & 49.23\% & 5.83& 9.93 & 59.42\% & 5.04  &  8.66  &  51.13\%\\
MIP & \uline{4.74}         & \uline{8.13} & \textbf{45.10\%} & \uline{4.56}  & \textbf{7.27} & \textbf{43.32\%} & \textbf{5.26}& \textbf{9.18} & \textbf{55.27\%} &    \textbf{4.87}  &  \textbf{8.16}  &  \textbf{47.56\%}\\
\midrule  
\multirow{2}{*}[-2pt]{NYCBike(Out)}            & \multicolumn{3}{c|}{testing set 0}                & \multicolumn{3}{c|}{testing set 1}                & \multicolumn{3}{c|}{testing set 2}      & \multicolumn{3}{c}{overall results}           \\
\cmidrule{2-4} \cmidrule{5-7} \cmidrule{8-10}\cmidrule{11-13}
    & MAE   & RMSE  & MAPE    & MAE    & RMSE  & MAPE    & MAE   & RMSE  & MAPE  & MAE   & RMSE  & MAPE    \\
\midrule  
STGCN      & 5.04                  & 8.84 & 46.64\% &                   4.78&7.83 & 47.81\% &                  5.51 & 9.56 & 62.02\% & 5.11 & 8.74  &  52.16\% \\
DCRNN      & 5.38          &9.04  & 54.46\% &                 4.97  & 7.84 & 52.51\% &  5.78                 & 10.15 & 65.90\% & 5.38  &  9.01  & 57.62\%\\
STNorm     & 5.19          & 9.23 & 44.80\% & 4.83         & 7.90 & \uline{43.63\%} & 5.57  & 9.70 & 58.47\% & 5.20  &  8.94  & 48.97\% \\
GMSDR      & 4.98                  & 8.75 & 45.79\% & \uline{4.73}                  & \uline{7.74} & 44.74\% & 5.55                  & \uline{9.56} & 60.67\% & \uline{5.09}  &  \uline{8.68}  &  50.40\% \\
MegaCRN    & 5.53                  & 9.58 & 49.97\% & 5.13                  & 8.49 & 48.76\% & 5.82                  & 10.48 & 64.23\%  &  5.49  &  9.51  &  54.32\% \\
CauSTG      & 5.04         & 8.73 & 46.13\% & 4.95  & 8.04 & 46.68\% & 5.60& 9.71 & \textbf{56.65\%} & 5.20  &  8.83  & 49.82\%  \\
TESTAM     & \textbf{4.88}         & \textbf{8.45} & \textbf{43.63}\% & 5.08  & 8.59 & 46.03\% & \uline{5.54}& 9.59 & \uline{56.81\%}  &  5.16  &  8.88 & \uline{48.82\%} \\
MIP & \uline{4.94}         & \uline{8.68} & \uline{44.72\%} & \textbf{4.66}  & \textbf{7.56} & \textbf{43.22\%} & \textbf{5.51}& \textbf{9.45} & 57.60\%  &  \textbf{5.03}  &  \textbf{8.56}  & \textbf{48.51\%}\\
\bottomrule
\end{tabular}
}
\vspace{-2mm}
\label{tab:results1}
\end{table*}

\noindent\textbf{Temporal Transformer Layer.}
Once the GNN layer processes all graphs at all time steps, we can collect $T$ feature matrices produced by the final graph propagation layer, denoted by $\mathbf{G}^1,\mathbf{G}^2,...,\mathbf{G}^T\in \mathbb{R}^{N\times d}$. For a certain node $n$, we can stack all its $d$-dimensional, time-sensitive features across $T$ steps into a matrix, denoted by $\mathbf{G}_n \in \mathbb{R}^{T \times d}$. With that, we learn the dependencies across all temporal features of a node through a transformer layer as:
\begin{equation}
\begin{aligned}
    \mathbf{G}_{n}' &= \text{softmax}\left(\frac{\mathbf{G}_{n}   \mathbf{W}_Q   \left(\mathbf{G}_{n}   \mathbf{W}_K \right)^{\top}}{\sqrt{d}} \right) \left(\mathbf{G}_{n}  \mathbf{W}_V \right),\\
    \mathbf{Z}_n &= \text{MLP}({\mathbf{G}}_{n}'),
\end{aligned}
\end{equation}
where $\mathbf{W}_Q, \mathbf{W}_K, \mathbf{W}_V \in \mathbb{R}^{d \times d}$ are trainable query, key and value projection weights, and MLP denotes a feedforward multilayer perceptron. 

\noindent\textbf{Prediction Layer.}
After obtaining $N$ outputs for all nodes $\mathbf{H}_1', \mathbf{H}_2', ..., \mathbf{H}_N' \in \mathbb{R}^{T\times d}$, we can stack all feature matrices into $\mathbf{Z} \in \mathbb{R}^{T\times N \times d}$. Then, we generate the final predictions with an MLP: 
\begin{equation}
    \mathbf{\hat{Y}} = \text{MLP}\left( \mathbf{Z'} \right),
\end{equation}
where the MLP projects $\mathbf{Z}$ into  $\mathbf{\hat{Y}} \in \mathbb{R}^{T \times N\times k}$ that carries the predicted urban flow per time step per location. 


\subsection{Model Optimization}
\label{framework}
Now, we detail the optimization strategy for MIP. Firstly, based on the prediction $\mathbf{\hat{Y}}\in \mathbb{R}^{L\times N\times k}$ generated by the backbone model, the prediction error is quantified as follows:
\begin{equation}
    \mathcal{L}_{task} = \frac{1}{TN} \sum_{t,n,k'=1}^{T,N,k}|\mathbf{\hat{Y}}[t,n,k']-\mathbf{Y}[t,n,k']|.
    \label{eq:taskloss}
\end{equation}

In addition, as we extract invariant prompts from a memory bank, we use an auxiliary regularization loss to enhance the quality of features stored within the memory bank:
\begin{equation}
\begin{aligned}
    &\mathcal{L}_{reg}=\sum_{t,n}^{T,N} \max \left\{\Vert \mathbf{H}_{I}^t[n]-\mathbf{\Phi}[a]\Vert^2 
    - \Vert \mathbf{H}_{I}^t[n]-\mathbf{\Phi}[b]\Vert^2 \right. \\
    &\quad \left. + \kappa, 0 \right\} + \sum_{t,n}^{T,N}\Vert \mathbf{H}_{I}^t[n]-\mathbf{\Phi}[a]\Vert^2,
\end{aligned}
\label{eq:auxiliaryloss}
\end{equation}
where $\kappa$ is a distance margin, $a,b$ are the indices of the most and second similar virtual nodes w.r.t. node $n$ based on the similarity score $\mathbf{S}^t_I$ computed in Eq.\ref{eq:Hinv}. Intuitively, the first term encourages the top two queried memory bank features to encode different information, while the second term aims to better align node $n$'s invariant feature with the top feature queried from the memory bank. As such, the combination of these two regularization terms improves the diversity and representativeness of all $M$ prototype features in the memory bank, thus refining the learned invariant features of each node at different time steps. 
Finally, the optimization objective aims to minimize the following overall loss:
\begin{equation}
    \mathcal{L}=\mathcal{L}_{task}+ \mathcal{L}_{inv}+\lambda_2  \mathcal{L}_{reg},
\label{eq:final_loss}
\end{equation}
with a balancing hyperparameter $\lambda_2$. As MIP is being trained towards convergence, the intervention on variant patterns, i.e., $\hat{\mathbf{H}}_V = \text{Intervene}(\mathbf{H}_V, r)$ is re-executed in every training epoch, so as to inject more variations in the supervision signals. It is worth noting that, once MIP is trained, only the spatial-temporal backbone model described in Section \ref{backbone} is activated for making predictions in the inference stage. 

\begin{figure*}[htbp]
\centering
    \begin{subfigure}{0.3\linewidth}
        \centering
        \includegraphics[width=\textwidth]{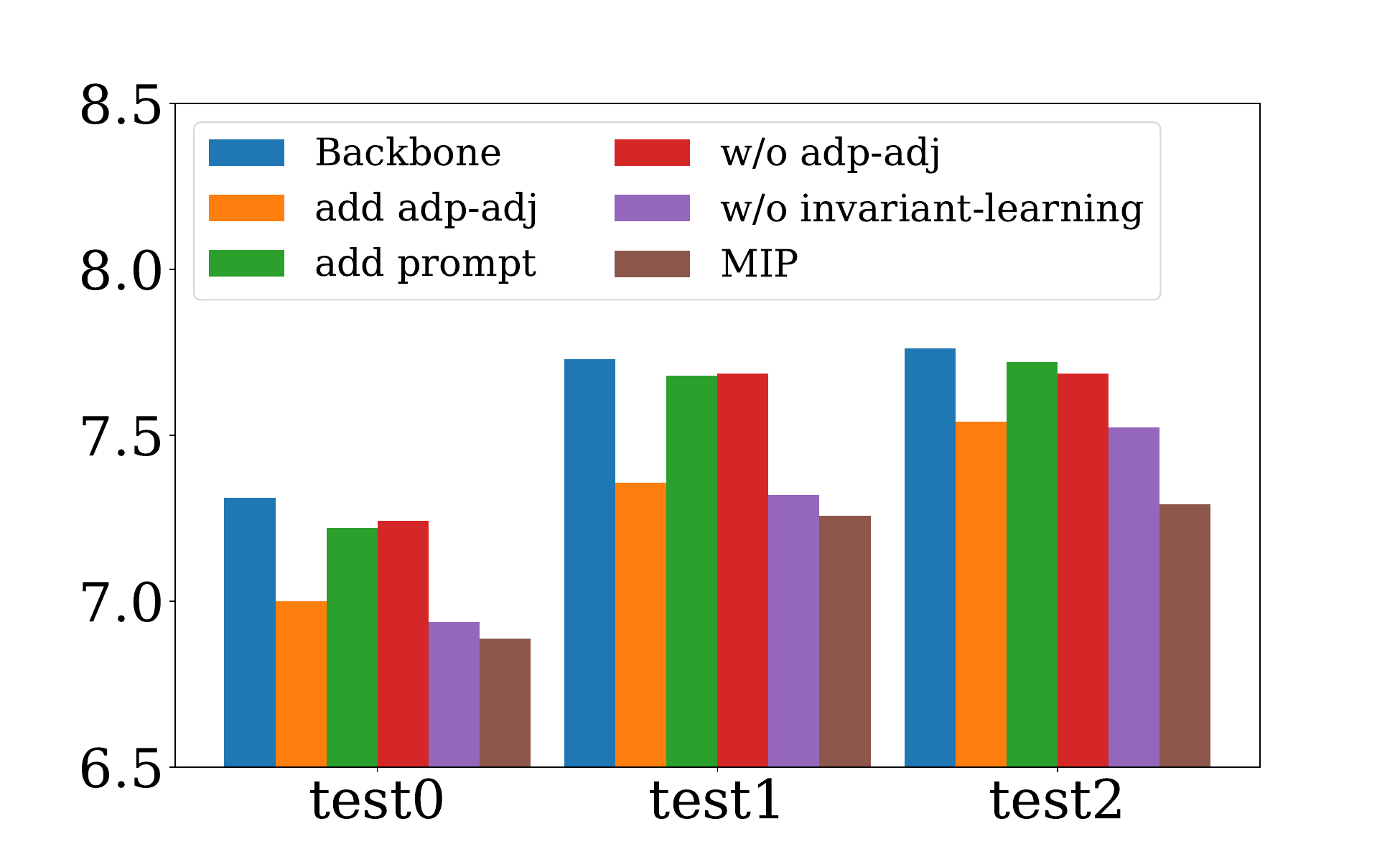}
        \caption{METR-LA dataset.}
        \label{fig:ablation_1}
    \end{subfigure}%
    \begin{subfigure}{0.3\linewidth}
        \centering
        \includegraphics[width=\textwidth]{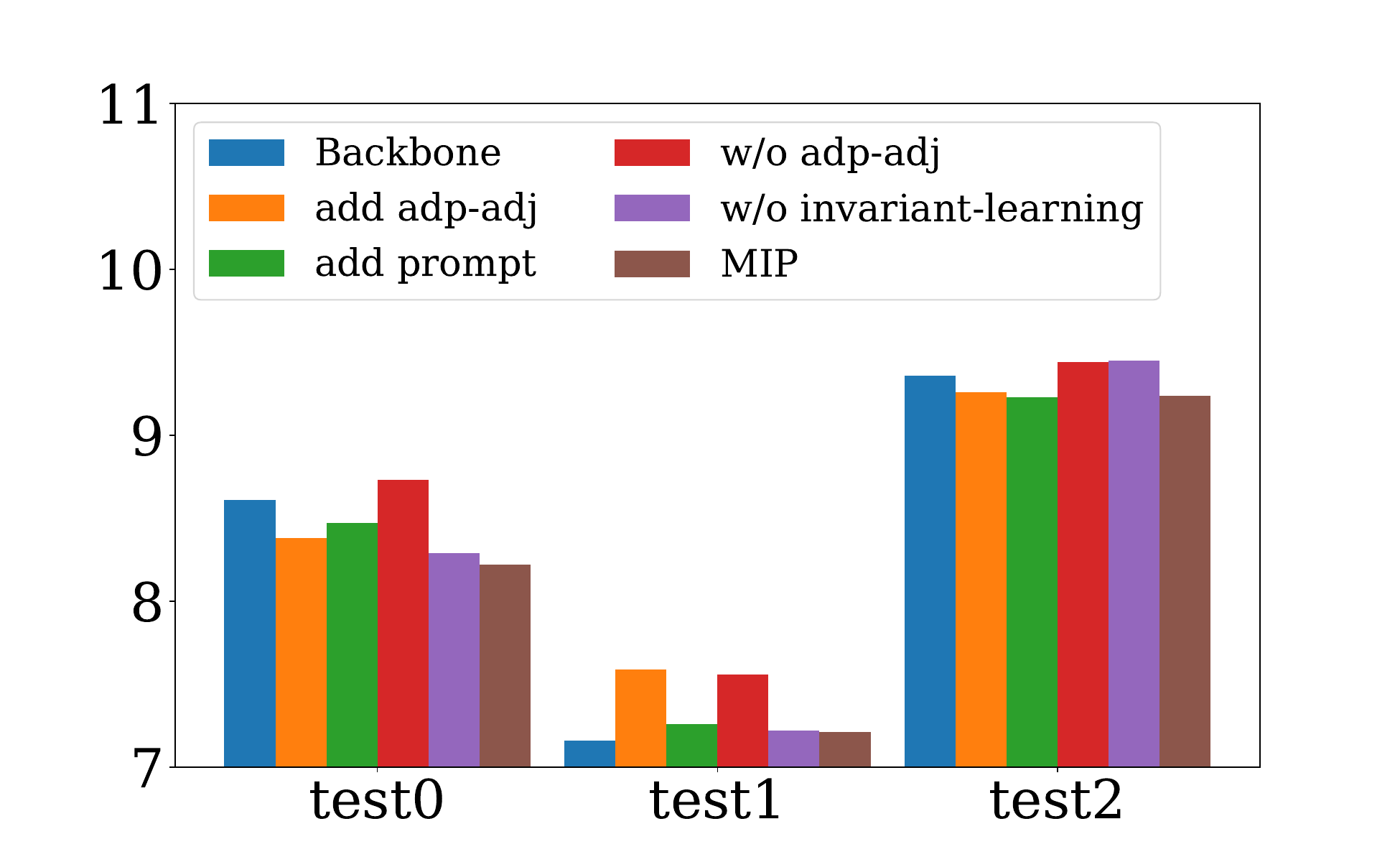}
        \caption{NYCBike1 dataset(In).}
        \label{fig:ablation_2}
    \end{subfigure}
    \begin{subfigure}{0.3\linewidth}
        \centering
        \includegraphics[width=\textwidth]{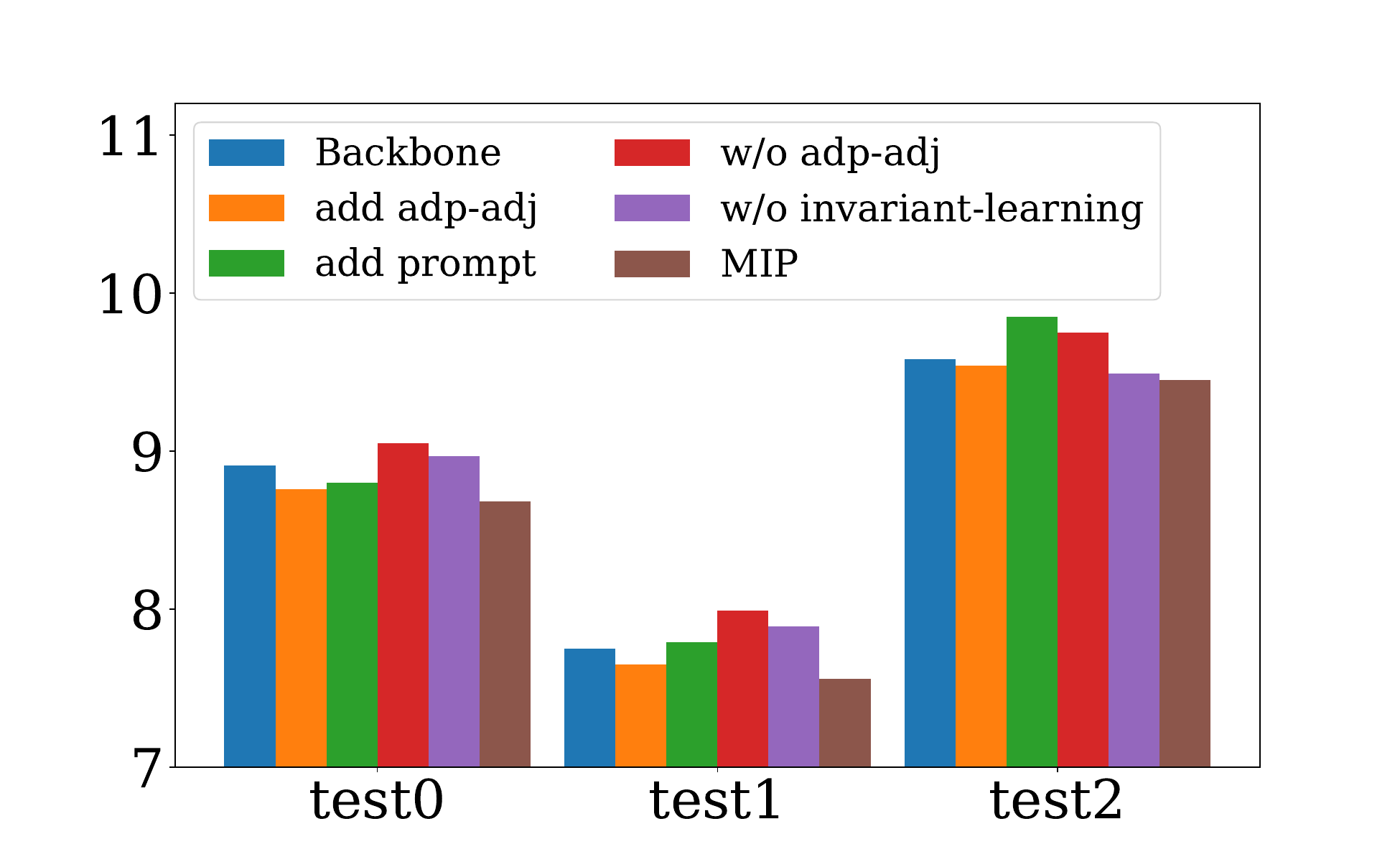}
        \caption{NYCBike1 dataset(Out).}
        \label{fig:ablation_3}
    \end{subfigure}
    \caption{Ablation study: RMSE of MIP and its variants.}
    \label{fig:ablation}
    \vspace{-5mm}
\centering
\end{figure*}

\begin{figure*}[htbp]
\centering
    \begin{subfigure}{0.3\linewidth}
        \centering
        \includegraphics[width=\textwidth]{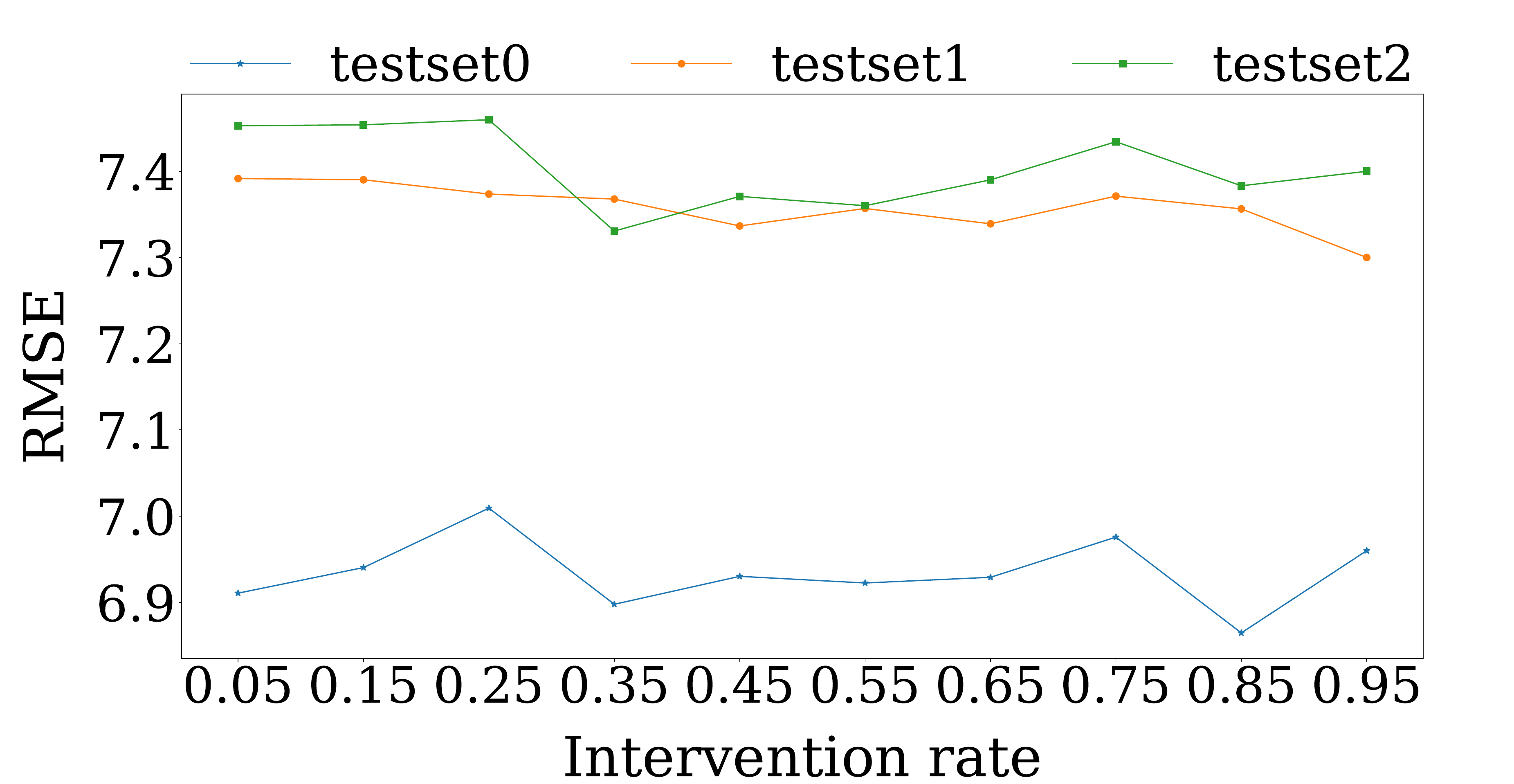}
        \caption{METR-LA dataset.}
        \label{fig:interventioinrate_1}
    \end{subfigure}%
    \begin{subfigure}{0.3\linewidth}
        \centering
        \includegraphics[width=\textwidth]{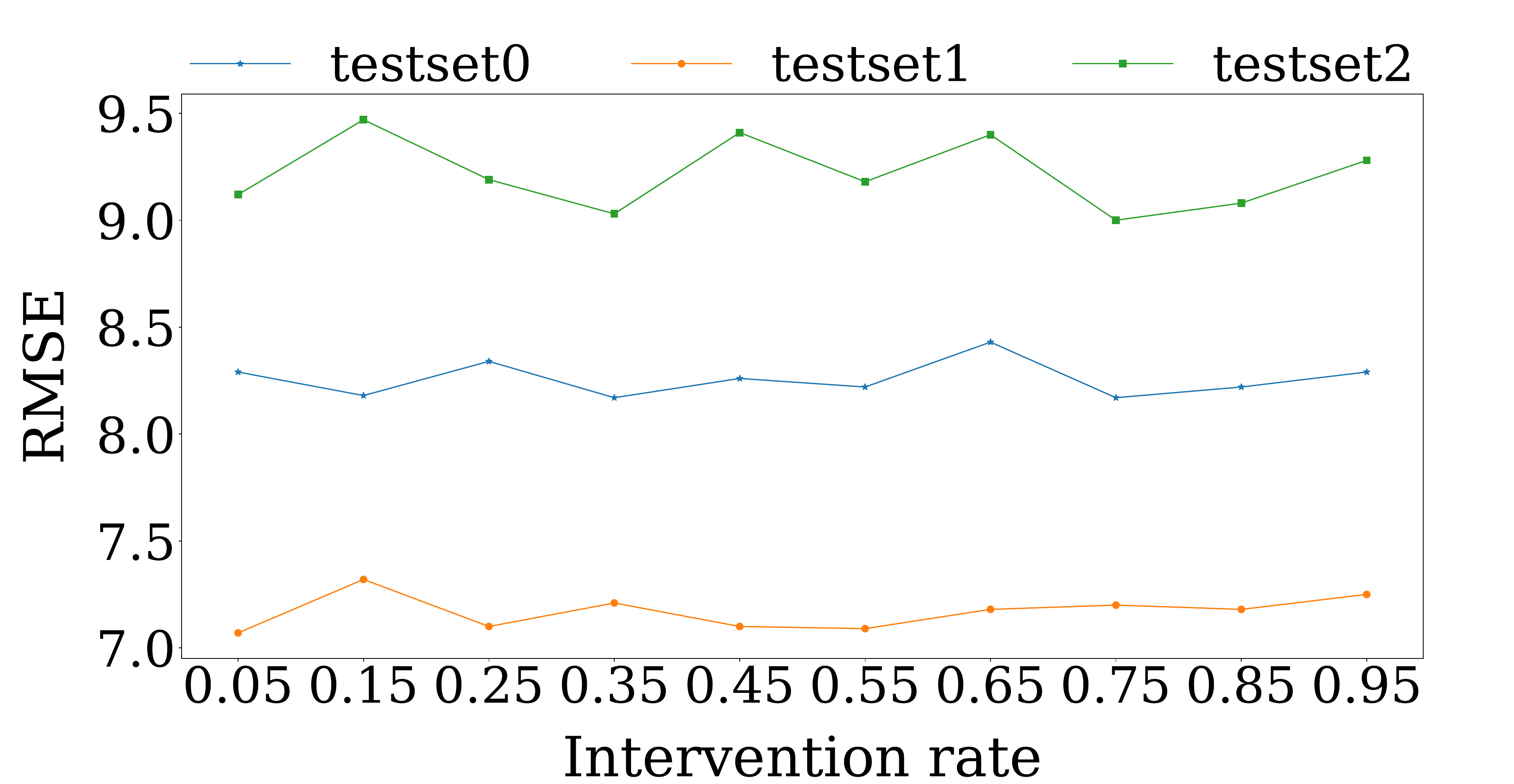}
        \caption{NYCBike1 dataset(In).}
        \label{fig:interventioinrate_2}
    \end{subfigure}
    \begin{subfigure}{0.3\linewidth}
        \centering
        \includegraphics[width=\textwidth]{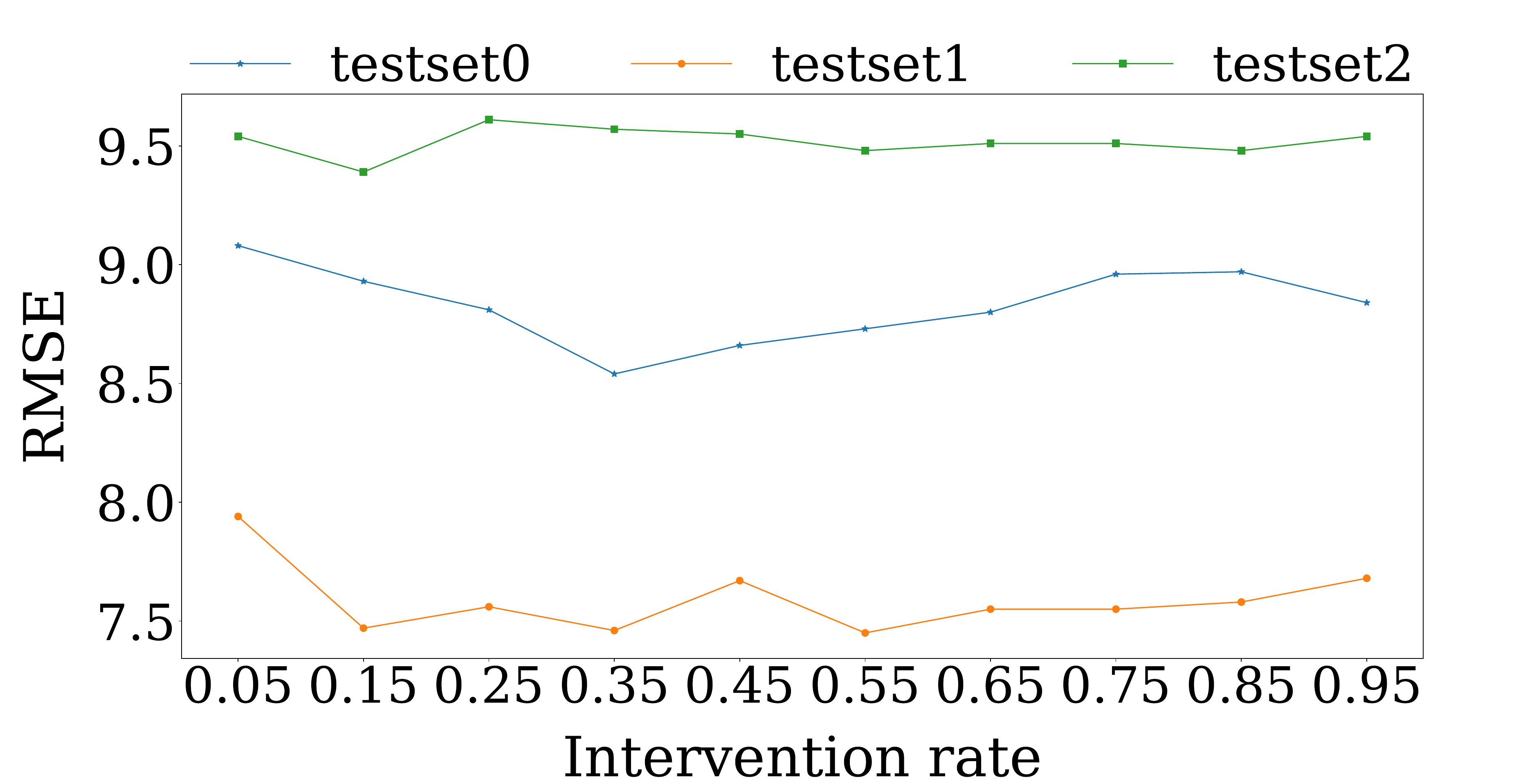}
        \caption{NYCBike1 dataset(Out).}
        \label{fig:interventioinrate_3}
    \end{subfigure}
    \caption{RMSE of MIP with different intervention rate.}
    \label{fig:intervention}
    \vspace{-5mm}
\centering
\end{figure*}

\subsection{Discussions}
\noindent\textbf{Inference Time Complexity Analysis.} 
Although MIP trains two models during the training phase, it exclusively utilizes the prediction model during inference, enabling it to deliver precise predictions with reduced time consumption. Thus, the time consumption in the inference stage is much smaller than in the training stage. The time complexity of extracting invariant prompt is $O(TNMd)$. The time complexity of one GNN layer is $O(ZTN^2d+ZTNd^2)$, where $Z$ is the order of information propagation. The time complexity of the temporal multi-head attention layer is $O(NT^2d+NTd^2)$. As we will demonstrate in Section \ref{sec:scalability}, MIP's efficiency during inference allows it to scale up to real-world datasets with varying numbers of locations and time steps. 

\noindent\textbf{Difference with Other Invariant Learning Methods.} 
With some simplifications of the preliminaries provided in Section \ref{ilearning}, OOD-robust models aim to learn $P(\mathbf{Y}|\mathbf{X}_s, E=E_s)$ instead of $P(\mathbf{Y}|\mathbf{X})$ to address the (OOD) problem, where $\mathbf{X}_s$ indicates the intervened input features under environment $E_s$. 
By using Bayesian rules, the posterior probability can be decomposed as:
\begin{equation}
    P(\mathbf{Y}|\mathbf{X}_s)=\sum_{s\in S}P(\mathbf{Y}\mid \mathbf{X},E=E_s)P(E=E_s).
\end{equation}
Linking back to the discussions earlier, most existing studies~\cite{yang2022towards,wu2022discovering,xia2024deciphering,wu2024graph} aim to establish models to approximate the right-hand side of the above equation. As a result, they need to enumerate the environments from which the samples were taken. This inevitably introduces additional parameterization for the environments, and the size of the underlying environments (i.e., $|\mathcal{S}|$) is highly dependent on the specific dataset used. In this work, by generating intervened variant prompts $\hat{\mathbf{H}}_V$ and substituting $\mathbf{X}_s$ with the concatenation of $\mathbf{H}_I$ and $\hat{\mathbf{H}}_V$, our model aims to directly learn the left-hand side of the equation, thus bypassing the need for designing sophisticated modules for environment modeling. 

\section{Experiments}
\label{exp}
In this section, we experimentally evaluate our model's generalizability under OOD settings in urban flow prediction tasks. Specifically, we aim to answer the following research questions (RQs): \textbf{RQ1:} Can MIP yield state-of-the-art performance in urban flow prediction with distributional shifts? \textbf{RQ2:} What is the contribution of each core component of MIP? \textbf{RQ3:} What is the sensitivity of MIP to its key hyperparameters? \textbf{RQ4:} Is MIP capable of learning invariant patterns from urban flow data to ensure generalizability? \textbf{RQ5:} What is the scalability and efficiency of MIP?

\begin{figure*}[htbp]
\centering
    \begin{subfigure}{0.3\linewidth}
        \centering
        \includegraphics[width=\textwidth]{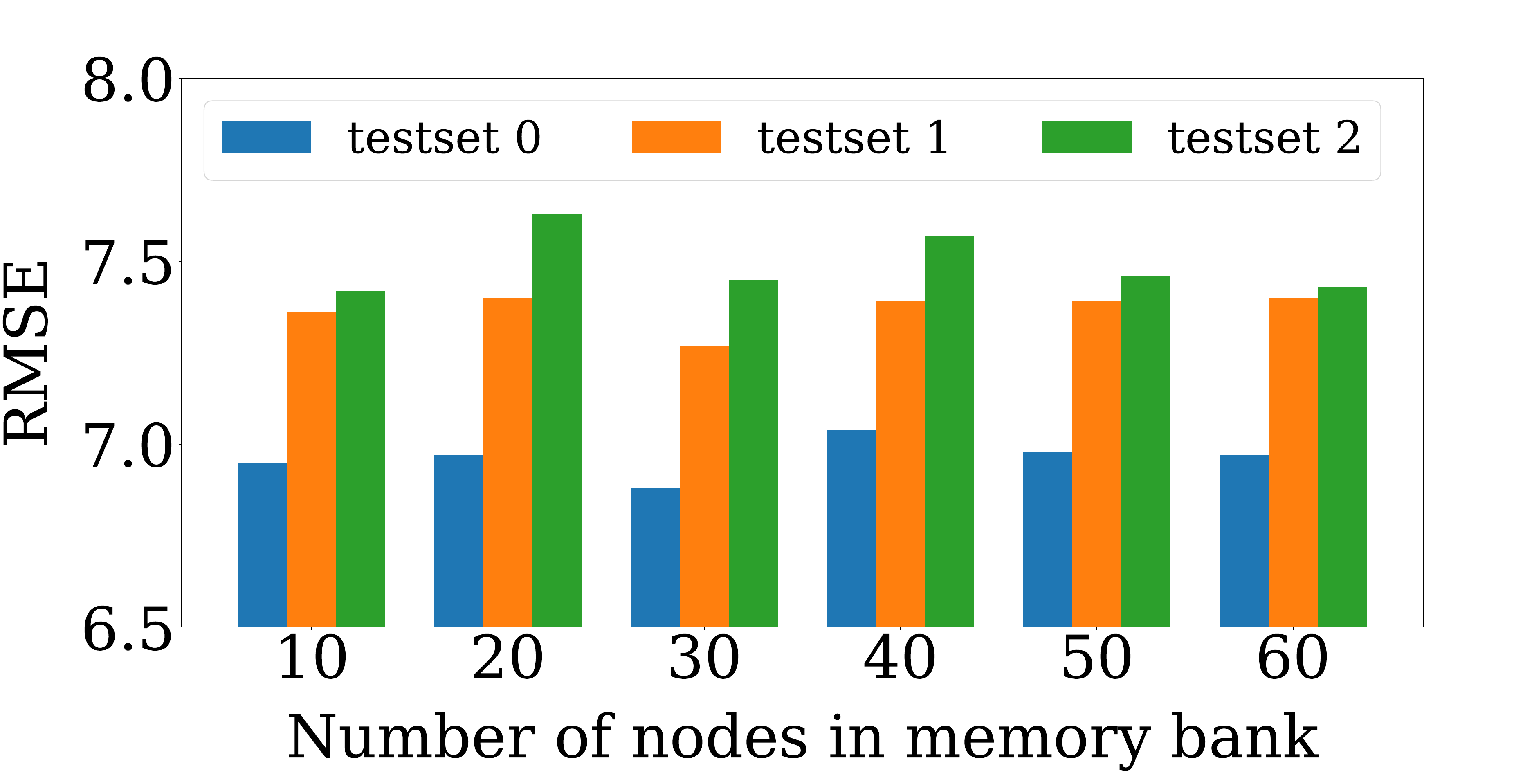}
        \caption{METR-LA dataset.}
        \label{fig:nodeinbank_1}
    \end{subfigure}%
    \begin{subfigure}{0.3\linewidth}
        \centering
        \includegraphics[width=\textwidth]{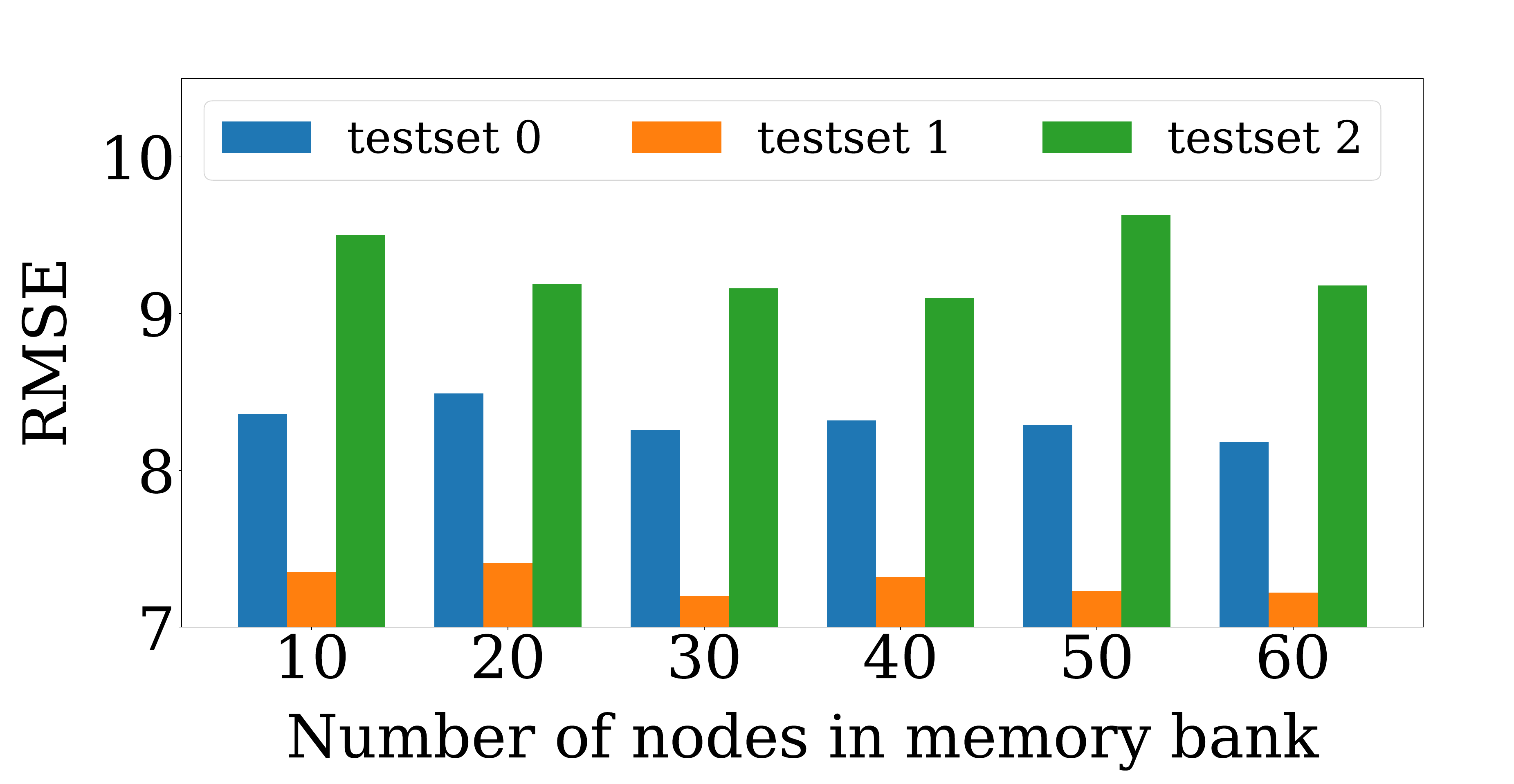}
        \caption{NYCBike1 dataset(In).}
        \label{fig:nodeinbank_2}
    \end{subfigure}
    \begin{subfigure}{0.3\linewidth}
        \centering
        \includegraphics[width=\textwidth]{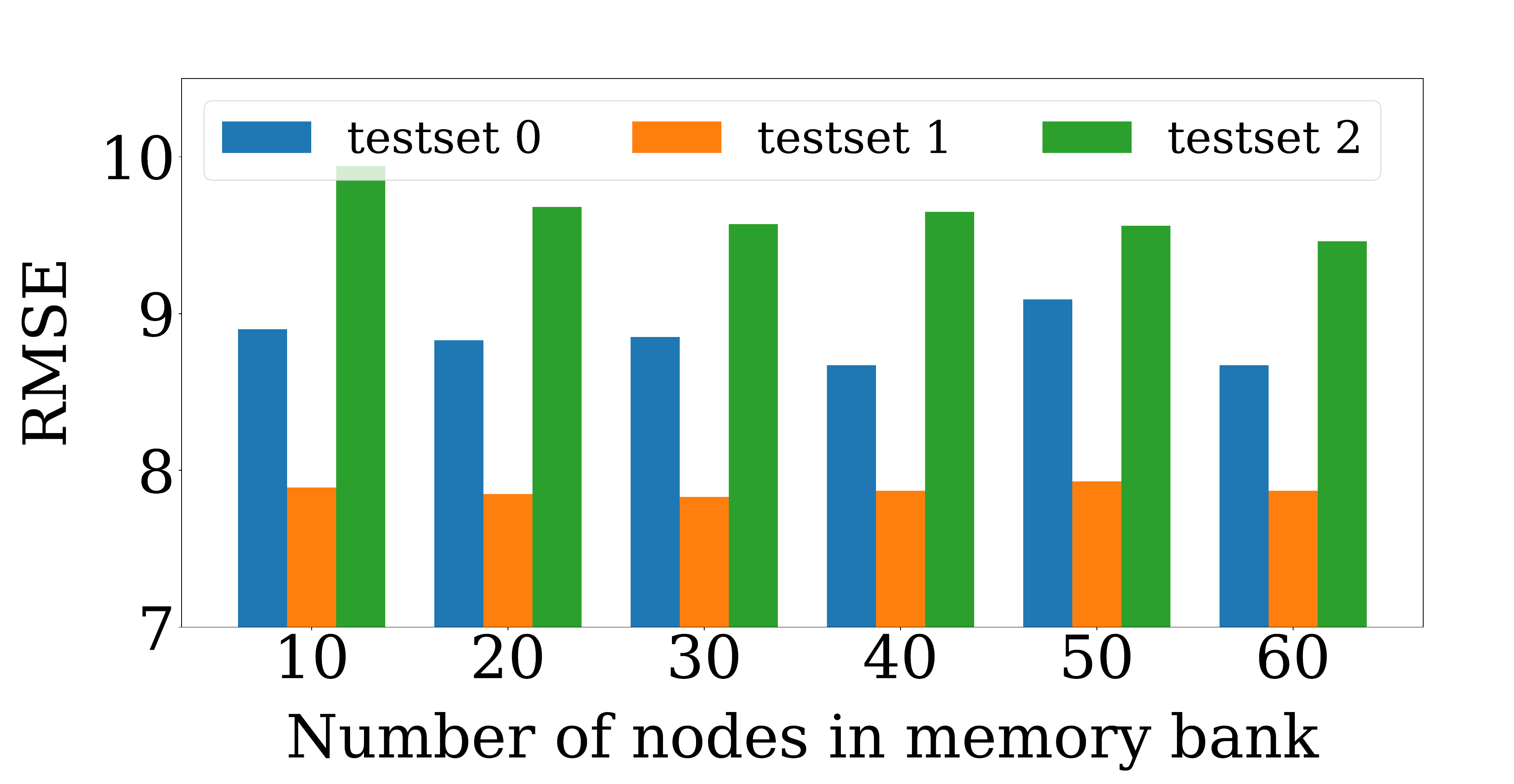}
        \caption{NYCBike1 dataset(Out).}
        \label{fig:nodeinbank_3}
    \end{subfigure}
    \caption{RMSE of MIP with a different number of nodes in the memory bank.}
    \label{fig:nodeinbank}
    \vspace{-5mm}
\centering
\end{figure*}

\subsection{Experimental Settings}
\label{exp1}
\subsubsection{Datasets}
We evaluate our model on two well-established benchmarks, namely METR-LA~\cite{li2017diffusion} and NYCBike~\cite{zhang2017deep}. METR-LA~\cite{li2017diffusion} is a traffic speed prediction dataset collected with 207 sensors across Los Angeles. NYCBike1~\cite{zhang2017deep}  is a dataset of bike rental records in New York City, where the city is divided into $8\times 16$ equally-sized grids. A high-level summary of the datasets used is shown in Table~\ref{tab:data}. 

\subsubsection{Evaluation Protocol}
We split both datasets chronologically: the first 60$\%$ is for training, the following 10$\%$ for validation, and three test sets are constructed by evenly slicing the remaining data (10$\%$ for each). This is to fully mimic real-world application scenarios where a trained model is expected to provide predictions for multiple consecutive time periods with varying distributions. 
For convenience, we number the tree test sets with 0, 1, and 2. Generally, as test sets 0-2 become farther apart from the training set in time, their distribution shifts tend to become stronger. 
Based on the number of time steps available, we predict the next 12 time steps based on the past 12 on METR-LA and predict the next 6 time steps based on the past 6 on NYCBike. 
Similar to previous studies~\cite{wu2019graph,liu2022msdr,wang2024stone}, We evaluate all methods in terms of Mean Absolute Error (MAE), Root Mean Square Error (RMSE), and Mean Absolute Percentage Error (MAPE). For all three metrics, lower results indicate better performance.

\subsubsection{Baselines} 
We compare $MIP$ with the following state-of-the-art baselines:

\noindent\textbf{STGCN}~\cite{yu2017spatio}: It alternately stacks GNN and TCN layers as spatial-temporal modules.

\noindent\textbf{DCRNN}~\cite{li2017diffusion}: It replaces the matrix multiplication in GRU with a more performant graph diffusion convolution layer.

\noindent\textbf{STNorm}~\cite{deng2021st}: It proposes two normalization strategies to separately handle temporal and spatial signals.

\noindent\textbf{GMSDR}~\cite{liu2022msdr}: It captures long-range dependencies among time steps with a new RNN variant.

\noindent\textbf{MegaCRN}~\cite{jiang2023spatio}: It captures spatial-temporal patterns with a memory-augmented encoder-decoder architecture. 

\noindent\textbf{CauSTG}~\cite{zhou2023maintaining}: It trains models with different environments, and merges parameters with minimal variance for inference.

\noindent\textbf{TESTAM}~\cite{lee2024testam}: It is based on a mixture-of-expert structure, where each expert model uses a tailored adjacency matrix.

\subsection{Hyper-parameters and Implementation Notes}
When training our MIP, we set 30 nodes in the memory bank, and the dimension of them is set as 32. Moreover, the backbone model consists of three spatial-temporal layers, each of which is stacked with a GNN layer and a temporal Transformer layer. On the METR-LA dataset, the $\lambda_1$ and $\lambda_2$ are set as 0.3 and 0.1, respectively. On the NYCBike1 dataset, the $\lambda_1$ and $\lambda_2$ are set as 0.1 and 0.01, respectively. To optimize the trainable parameters, the Adam optimizer is used with learning rates at 0.001 and 0.003 on the two datasets. Additionally, the batch size for the METR-LA dataset is set to 64, while for the NYCBike1 dataset, it is set to 32. Besides, 25$\%$ of nodes are implemented intervention in both datasets. 

The codes are written in Python 3.8 and the operating system is Ubuntu 20.0. We use Pytorch 1.11.7 on CUDA 12.2 to train models on GPU. All experiments are conducted on a machine with 12th Gen Intel(R) Core(TM) i7-12700K CPU and NVIDIA RTX A5500 GPU with 24GB GPU memory.

\begin{figure*}[htbp]
\centering
    \begin{subfigure}{0.3\linewidth}
        \centering
        \includegraphics[width=\textwidth]{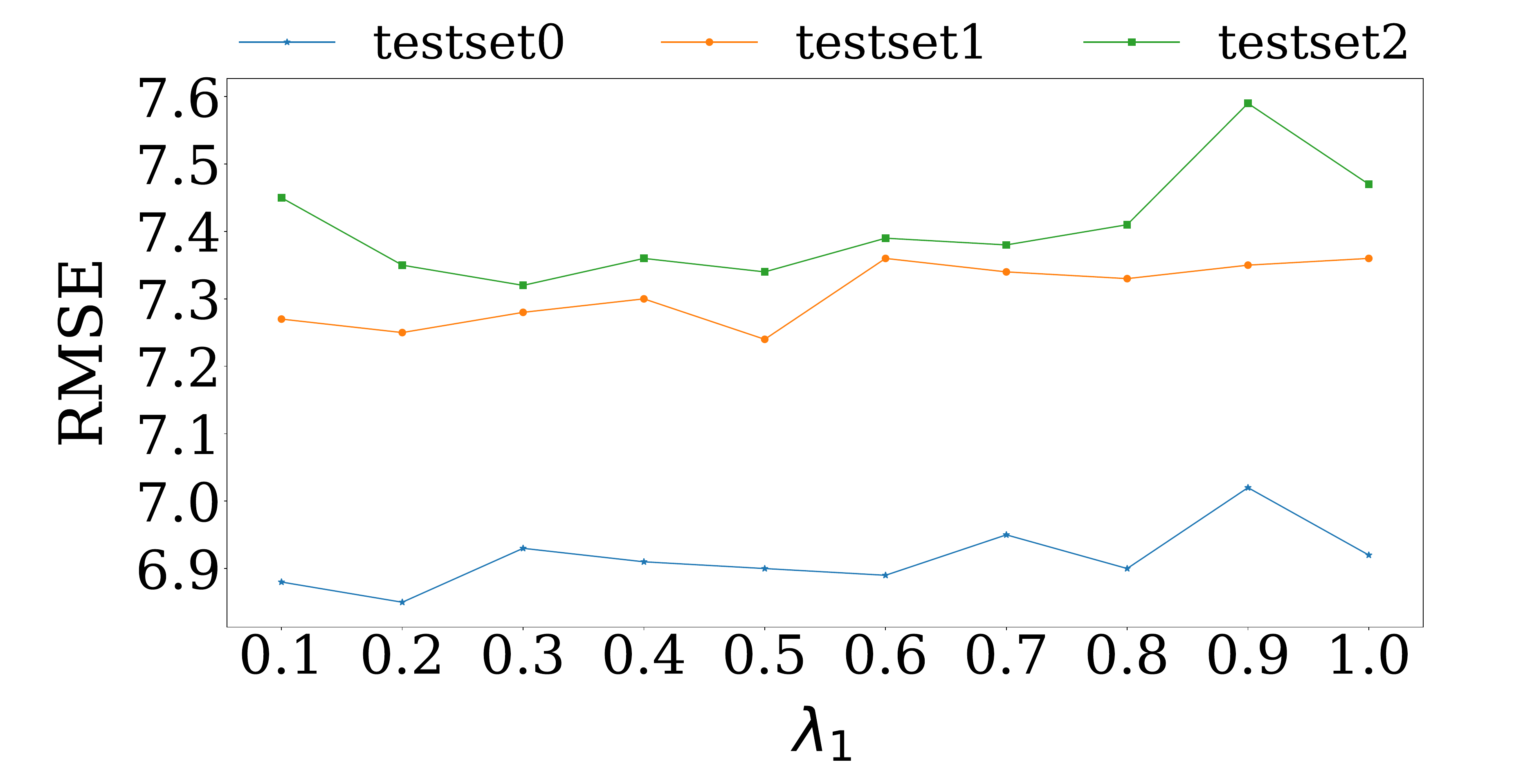}
        \caption{METR-LA dataset.}
        \label{fig:lamada1_1}
    \end{subfigure}%
    \begin{subfigure}{0.3\linewidth}
        \centering
        \includegraphics[width=\textwidth]{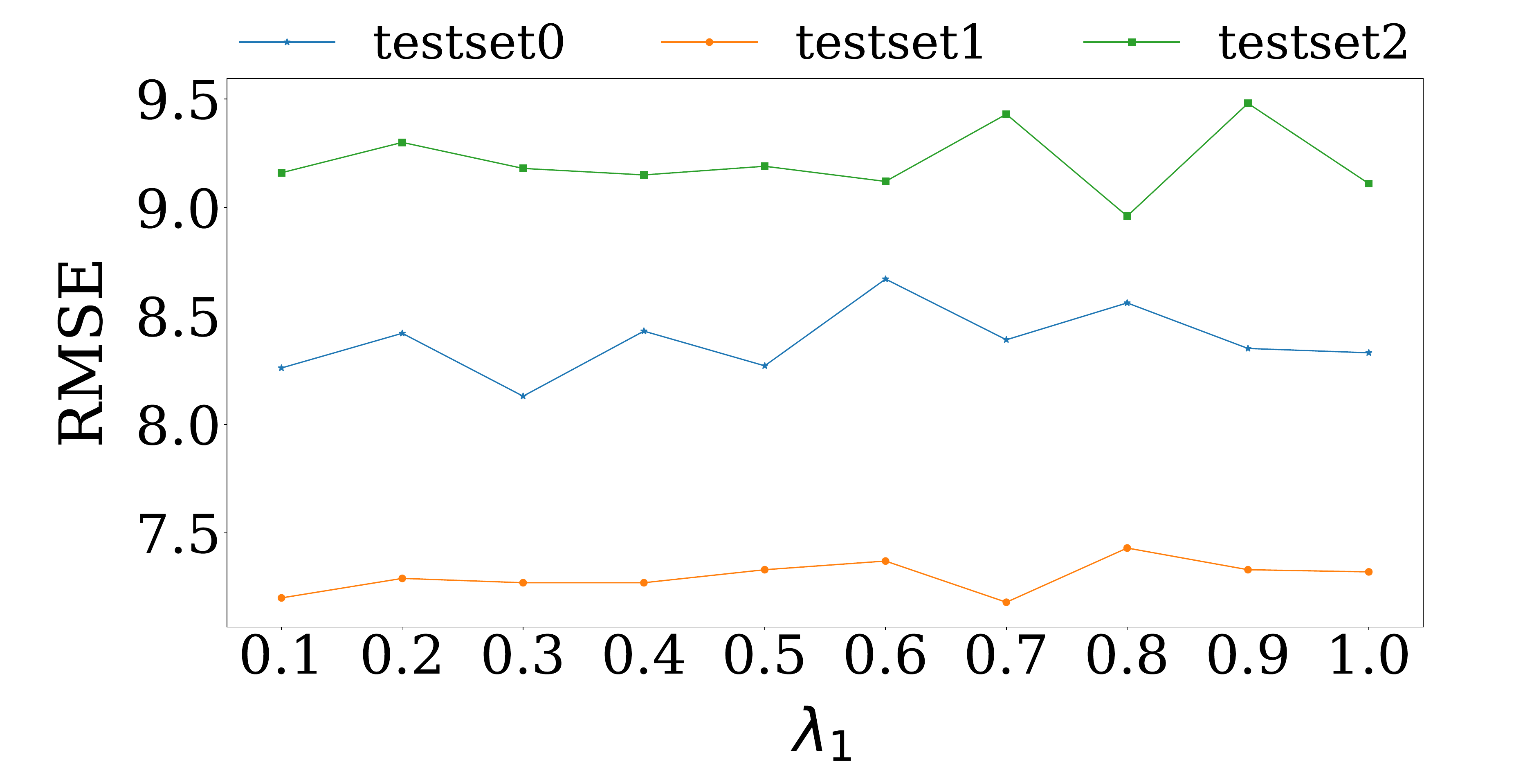}
        \caption{NYCBike1 dataset(In).}
        \label{fig:lamada1_2}
    \end{subfigure}
    \begin{subfigure}{0.3\linewidth}
        \centering
        \includegraphics[width=\textwidth]{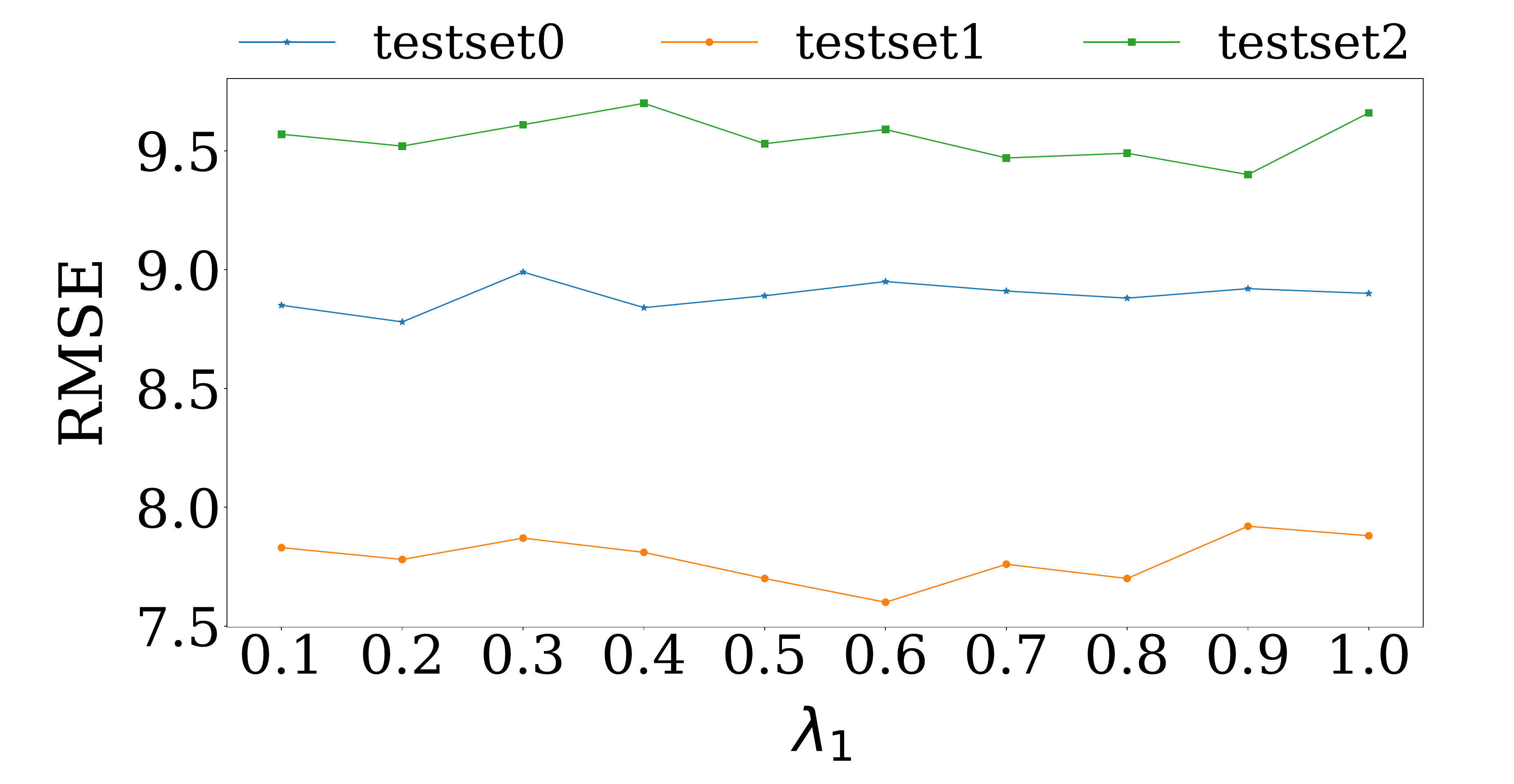}
        \caption{NYCBike1 dataset(Out).}
        \label{fig:lamada1_3}
    \end{subfigure}
    \caption{RMSE of MIP with different settings of $\lambda_1$.}
    \label{fig:lamada1}
    \vspace{-5mm}
\centering
\end{figure*}

\begin{figure*}[htbp]
\centering
    \begin{subfigure}{0.3\linewidth}
        \centering
        \includegraphics[width=\textwidth]{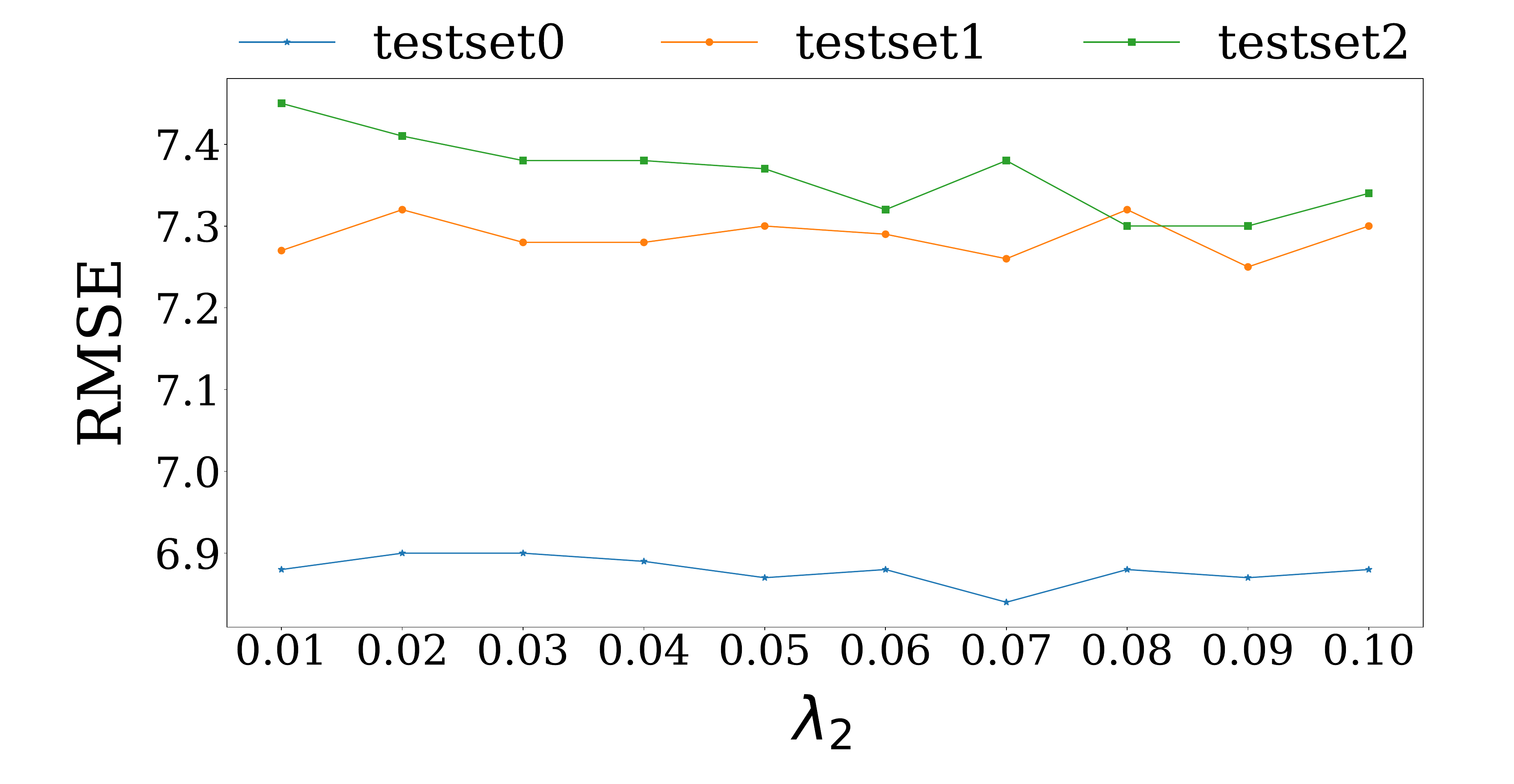}
        \caption{METR-LA dataset.}
        \label{fig:lamada2_1}
    \end{subfigure}%
    \begin{subfigure}{0.3\linewidth}
        \centering
        \includegraphics[width=\textwidth]{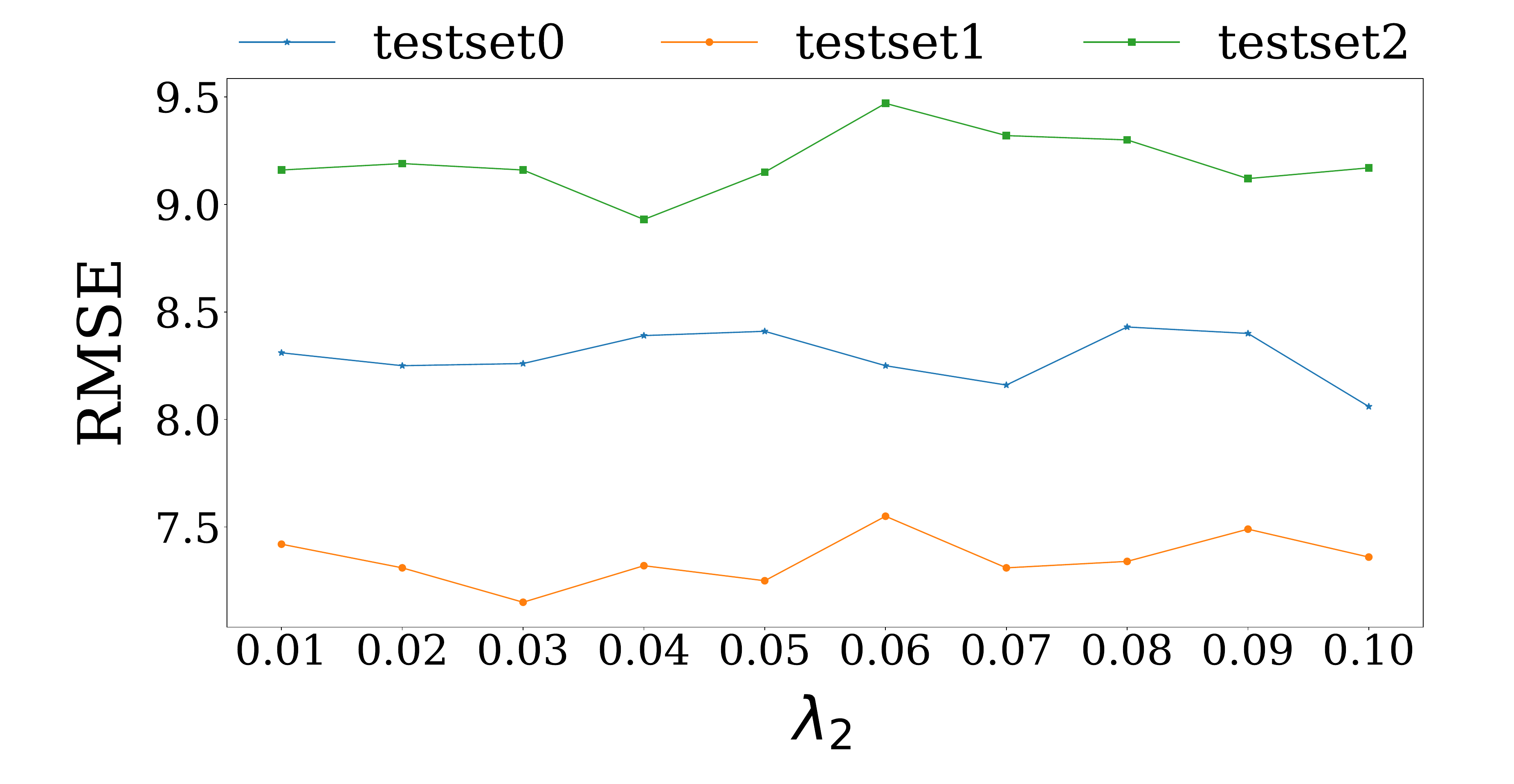}
        \caption{NYCBike1 dataset(In).}
        \label{fig:lamada2_2}
    \end{subfigure}
    \begin{subfigure}{0.3\linewidth}
        \centering
        \includegraphics[width=\textwidth]{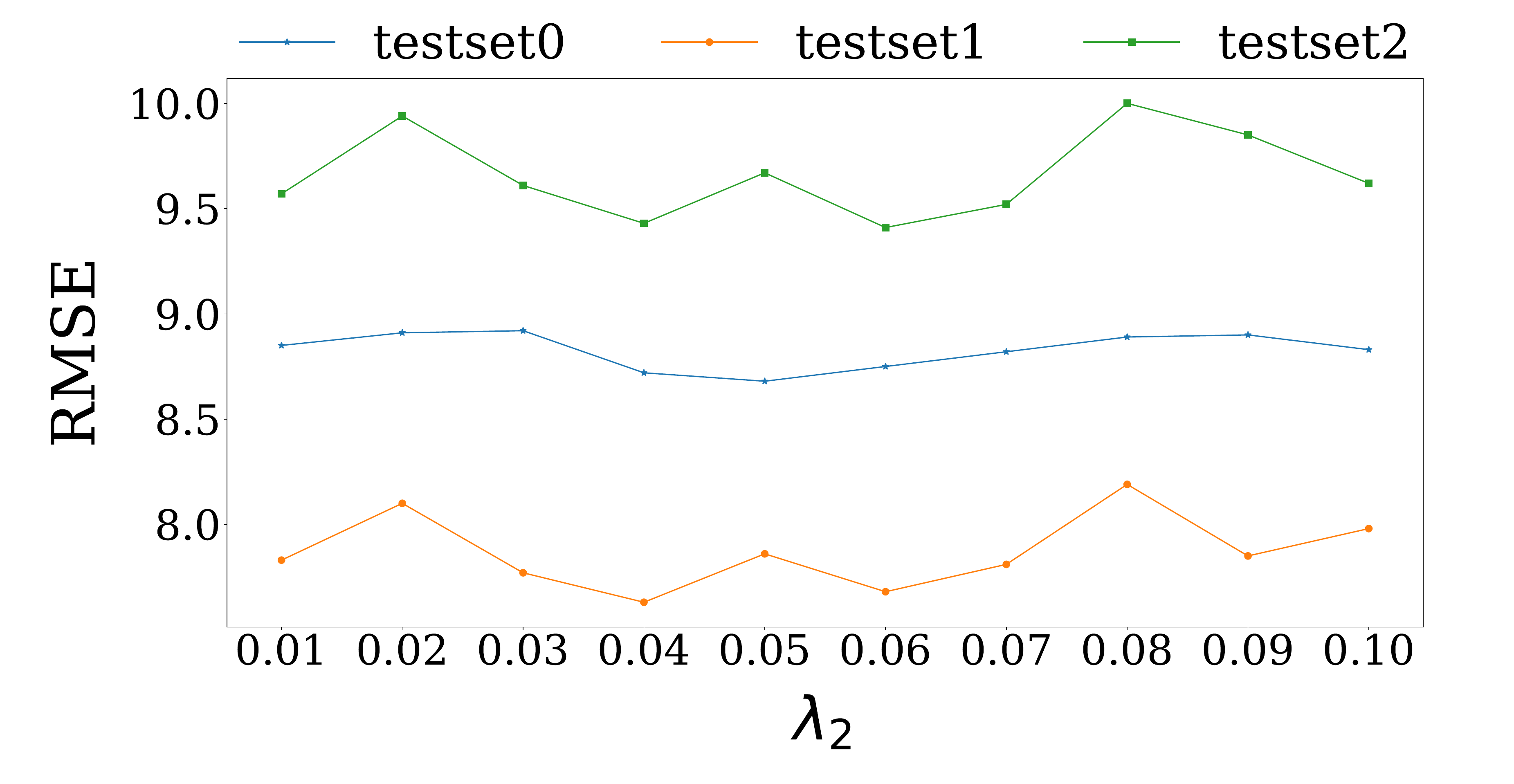}
        \caption{NYCBike1 dataset(Out).}
        \label{fig:lamada2_3}
    \end{subfigure}
    \caption{RMSE of MIP with different settings of $\lambda_2$.}
    \label{fig:lambda2}
    \vspace{-5mm}
\centering
\end{figure*}

\subsection{Performance Comparison with Baselines (RQ1)}
\label{exp2}
To answer RQ1, we compare our MIP with some SOTA baseline models, and the matrices of the final horizon are recorded in Table~\ref{tab:results1}. 
All in all, MIP clearly outperforms all competing baselines over the three testing sets, whereas the second-best performing model is not consistent across all cases. This indicates that MIP demonstrates high generalization ability across various datasets, highlighting its versatility and adaptability to various applications in urban flow prediction.
On the METE-LA dataset, as testing set 0 is the closest to the training dataset, its distribution changes less than the other two test sets. Thus, the models achieve similar performance on test 0, and some baselines get the best result, such as the MAPE of MegaCRN on testing set 0. However, the distribution shift happens more on test 1 and test 2, and the performance of all the models becomes worse. Some baseline models, such as MegaCRN, TESTAM, and GMSDR, get similar RMSE on testing set 0, while their RMSE on testing set 1 and testing set 2 increases largely. On the NYCBike1 dataset, all the models perform well on testing set 1 on both bikes' in and out tasks. A reasonable explanation is that the distribution of this testing set is more similar to that of the training set.
Although TESTAM achieved the best performance on all three evaluation matrices on testing set 0, it performed worse on testing sets 1 and 2, even with the biggest RMSE on testing set 1. 
Notably, while some models outperform our MIP on testing Set 0, they struggle with distribution shifts and consequently exhibit performance degradation on testing set 2. Moreover, when evaluating these models across all testing sets (referred to as overall results) our model consistently delivers superior performance across all evaluation metrics, with the sole exception of achieving second place in the Mean Absolute Error (MAE) for the METR-LA dataset. In conclusion, our model demonstrates the ability to provide stable predictions across testing sets that exhibit a variety of distribution shifts.

\subsection{Ablation Study (RQ2)}
To explore the significance of each core component in MIP and answer RQ2, we carry out an ablation study with the following variants:
\begin{itemize}
    \item \textbf{Backbone}: This variant is the backbone model alone.
    \item \textbf{add adp-adj}: Based on the backbone model, only the semantic adjacency matrix is added and the invariant learning loss is not used.
    \item \textbf{add prompt}: Only the prompt learned from the memory bank is fed into the backbone model, the semantic adjacency matrix and the invariant learning loss are not used.
    \item \textbf{w/o adp-adj}: This variant removes the adaptive adjacency matrix from MIP and remains the invariant learning loss.
    \item \textbf{w/o invariant learning}: In this variant, we remove the invariant loss but retain other components.
\end{itemize}

The results are presented in Fig.~\ref{fig:ablation}. We can see that MIP beats all the variants on both datasets. The backbone model gets the worst results on most of the test sets, as the naive GNNs and temporal Transformer layers cannot capture the distribution shift and the heterophily of the node features. 
\textbf{add adp-adj} perform much better than the backbone model, even better than \textbf{add prompt} and \textbf{w/o adp-adj} sometimes, as the semantic adjacency matrix connects nodes with similar urban flow data, even if they are far away from each other in topology. 
\textbf{add prompt} performs better than the backbone model, as it learns some useful features from the memory bank, while the improvement is tiny because the invariant and variant features are not separated for the lack of invariant learning. 
The performance of \textbf{w/o adp-adj} decreases much more than \textbf{MIP}. Even though it distinguishes the invariant and variant prompts with the invariant learning loss, its RMSE is bigger than the \textbf{w/o invariant learning}. As the invariant prompts are propagated to their distance-based neighbours rather than their semantic-based neighbours, some nodes receive the opposite features than their own features.
\textbf{w/o invariant learning} gets worse results than \textbf{MIP} as it mixes the invariant and variant prompts together for the absence of invariant learning. 

\subsection{Parameter Sensitivity(RQ3)}
To answer RQ3, we evaluate MIP with various hyper-parameters. 

\noindent\textbf{Intervention rate}: The intervention rate is closely related to the ability to separate the invariant and variant features. 
We set this parameter from 0.05 to 0.95 with an interval of 0.1, and evaluate our model on both datasets. 
In Fig.~\ref{fig:intervention}, we record the RMSE of the final horizon. The MIP demonstrates insensitivity to varying intervention rates, as evidenced by the small fluctuating RMSE across different levels of intervention. In the spatial-temporal model, the intervened variant prompts will propagate to all the nodes in an urban graph, even a small intervention rate will make all the nodes contain variant patterns before the prediction layer. Thus, the change in intervention rate does not influence the prediction RMSE. 

\noindent\textbf{Number of nodes in memory bank}:
We investigate the sensitivity of the number of nodes in the memory bank and show the results in Fig.~\ref{fig:nodeinbank}. The MIP performs well on all the datasets with 30 nodes in the memory bank. With a small number of nodes, MIP cannot extract high-quality invariant features due to the limited diversity. On the contrary, with more nodes in the memory bank, diverse invariant features lead to the increasing training difficulty of both the prediction model and invariant learning backbone model.

\noindent\textbf{The composition ratio of the loss function}: 
In Eq.~\ref{eq:final_loss}, the loss function consists of task loss, invariant loss, and auxiliary loss. The composition ratio of the last two losses is controlled with hyper-parameters $\lambda_1$ and $\lambda_2$, and we implement an experiment to investigate which one works more. Firstly, we set $\lambda_2=0.01$, and $\lambda_1$ from 0.1 to 1.0, with the step as 0.1, and record the RMSE of the last horizon on both datasets in Fig.~\ref{fig:lamada1}. On all the datasets, the RMSE of the testing set 0 is stable, and it fluctuates on testing sets 1 and 2. Concretely, on the METR-LA dataset, the model gets better generalization ability at $\lambda_1$ is 0.3, as the RMSE on the testing set 2 is the smallest. As for the NYCBike1 dataset, the RMSE also fluctuates on testing set 1 and testing set 2, which indicates that the MIP is not sensitive to this hyper-parameter on this dataset. 
Furthermore, we set $\lambda_1=0.1$ and $\lambda_2$ from 0.01 to 0.1, with the step as 0.01, and the results are shown in Fig.~\ref{fig:lambda2}. On all the datasets, the RMSE on test sets 0 and 1 fluctuates. On the METR-LA dataset, the RMSE on testing set 2 gradually decreases as $\lambda_2$ rises, as there are more nodes in this dataset, the proportion of $\mathcal{L}_1$ and $\mathcal{L}_2$ should be larger to make sure the invariant prompt are diversity enough for all the nodes. While on the NYCBike1 dataset, the RMSE on testing set 2 reaches the low point at about 0.04 or 0.05, as the number of nodes in this dataset is less than it in METR-LA, a low proportion of $\mathcal{L}_1$ and $\mathcal{L}_2$ can make the invariant prompt to be diversity enough for NYCBike1 dataset. All these experiments are carried out with 30 nodes in the memory bank. 

\begin{figure*}[htp]
\centering

\begin{subfigure}[b]{0.43\textwidth}
    \centering
    \includegraphics[width=\textwidth]{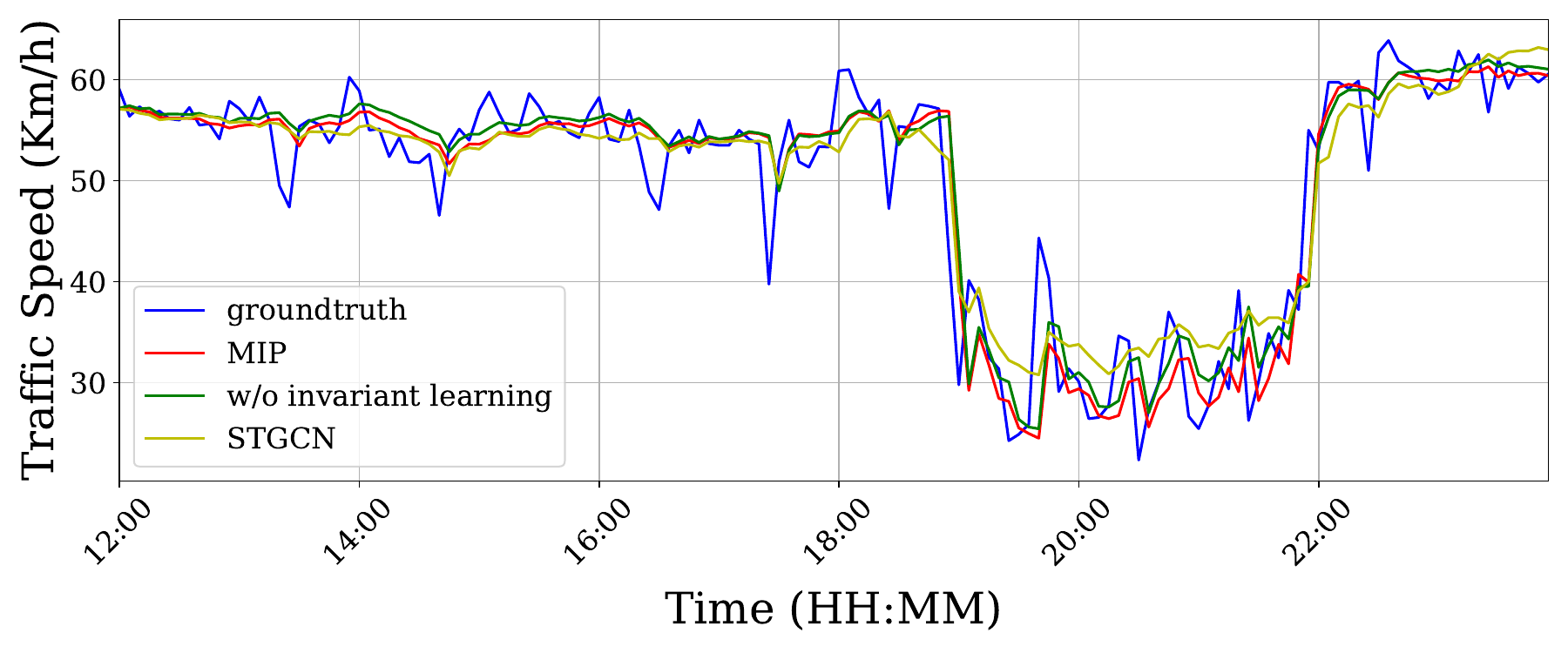}
    \caption{Predictions sampled from the first horizon.}
    \label{fig:casestudy1}
\end{subfigure}
\hspace{0.03\textwidth} 
\begin{subfigure}[b]{0.43\textwidth}
    \centering
    \includegraphics[width=\textwidth]{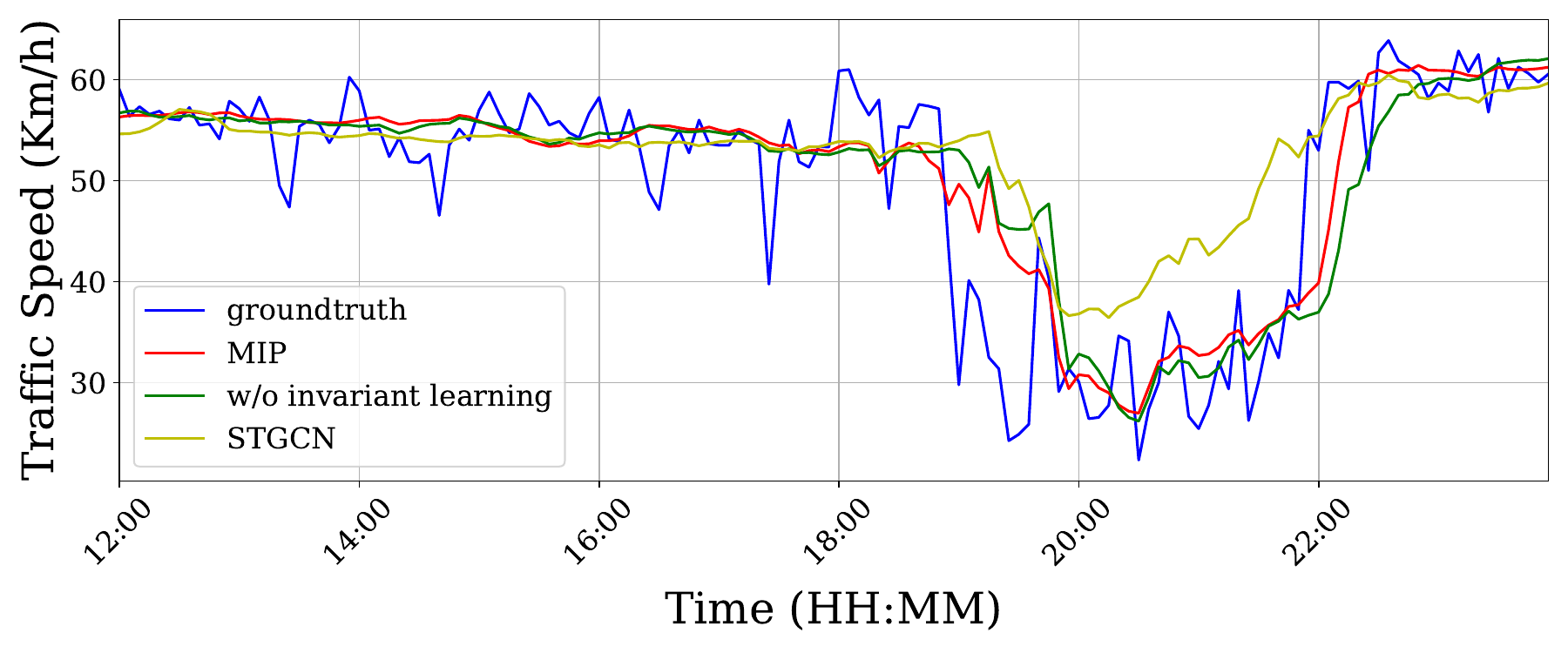}
    \caption{Predictions sampled from the eleventh horizon.}
    \label{fig:casestudy2}
\end{subfigure}

\vspace{-0cm} 

\begin{subfigure}[b]{0.95\textwidth}
    \centering
    \includegraphics[width=0.45\textwidth]{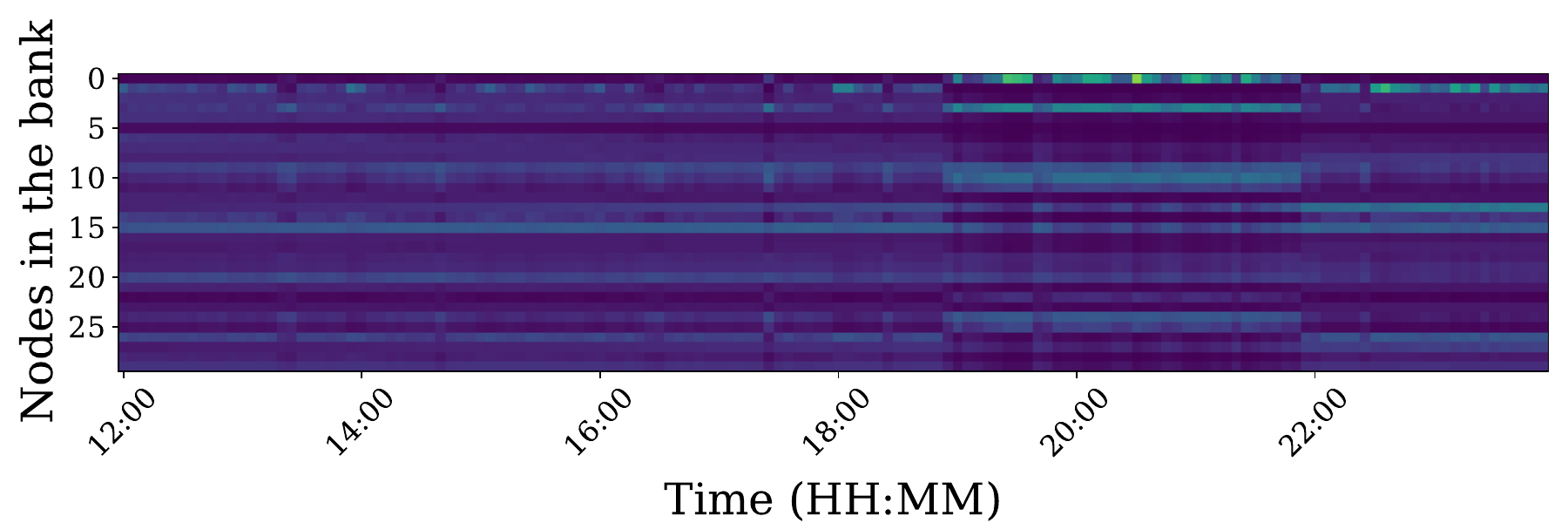}
    \hspace{0.03\textwidth}  
    \includegraphics[width=0.45\textwidth]{figures/horizon10_heatmap_without_invariant_learning.pdf}
    \caption{Prompt scores in the eleventh horizon from MIP, generated without (left) and with (right) invariant learning.}
    \label{fig:casestudy34}
\end{subfigure}

\caption{Case study on MIP's prediction performance under distribution shifts.}
\label{fig:casestudy}
\vspace{-5mm}
\end{figure*}

\subsection{Case Study (RQ4)}
We visualize the predictions and the prompt scores of a randomly selected node from the METR-LA dataset in Fig.~\ref{fig:casestudy}. In time series forecasting, distribution shifts often become more pronounced near the end of the prediction sequence, leading to increased prediction errors in many baseline models. In Fig.~\ref{fig:casestudy1} and Fig.~\ref{fig:casestudy2}, we present the prediction from the first and eleventh prediction time steps, respectively, of each sample used in the prediction task, which forecasts the next 12 time steps based on the preceding 12. In Fig.~\ref{fig:casestudy1}, the predictions from each model closely align with the ground truth, indicating their proficiency in short-term forecasting, likely due to minimal distribution differences between the first prediction time step and the preceding twelve time steps. 
In Fig.~\ref{fig:casestudy2}, all models exhibit significant deviations in their predictions compared to the ground truth. However, our model's predictions are closer to the ground truth, especially when the traffic speed decreases or increases dramatically at about 19:00 p.m. and 22:00 p.m. 
Moreover, we also show the prompt scores of the eleventh horizon of MIP and its variant \textit{w/o invariant learning} in Fig.~\ref{fig:casestudy34}. Notably, without the help of invariant learning, the variant \textit{w/o invariant learning}performs worse than MIP, and their prompt scores show different tendencies. This demonstrates invariant learning helps the model to extract invariant prompts and overcome the OOD problem.

\begin{figure}[b]
\centering
    \begin{subfigure}{0.45\linewidth}
        \centering
        \includegraphics[width=\textwidth]{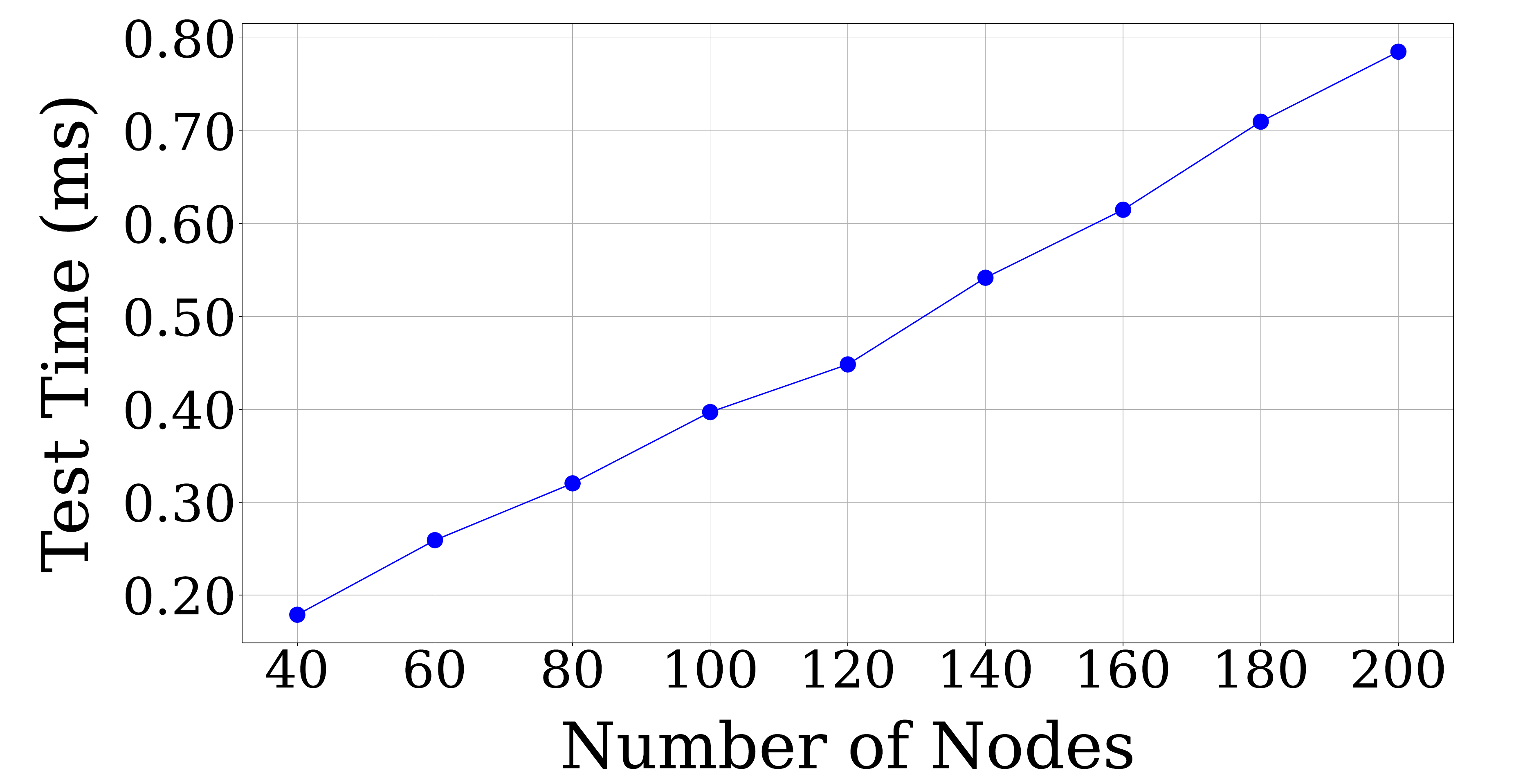}
        \caption{METR-LA dataset.}
        \label{fig:node_number_LA}
    \end{subfigure}%
    \begin{subfigure}{0.45\linewidth}
        \centering
        \includegraphics[width=\textwidth]{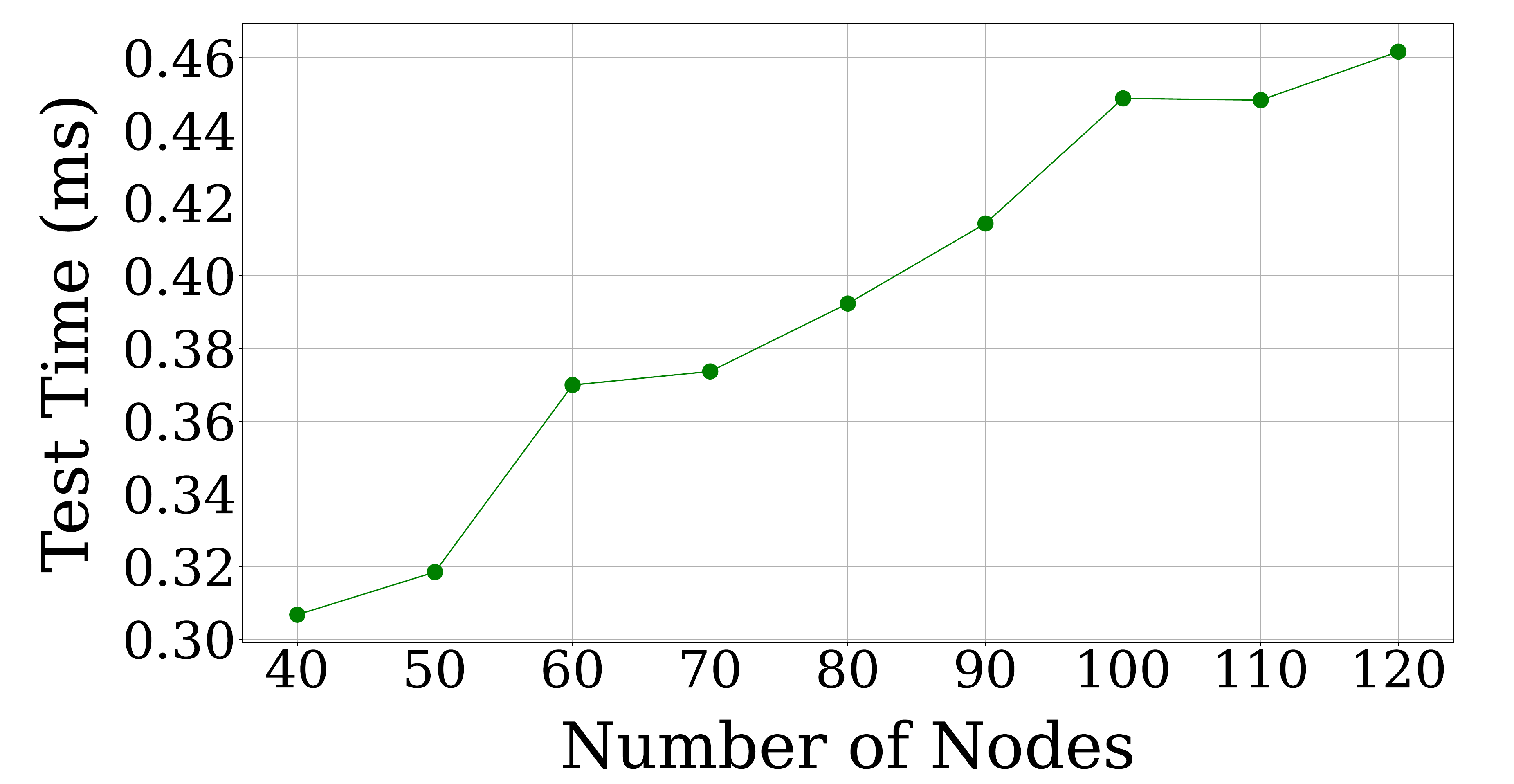}
        \caption{NYCBike1 dataset.}
        \label{fig:node_number_Bike}
    \end{subfigure}
    \caption{Inference time with different numbers of nodes $N$.}
    \label{fig:node_number}
    \vspace{-5mm}
\centering
\end{figure}
\begin{figure}[b]
\centering
    \begin{subfigure}{0.45\linewidth}
        \centering
        \includegraphics[width=\textwidth]{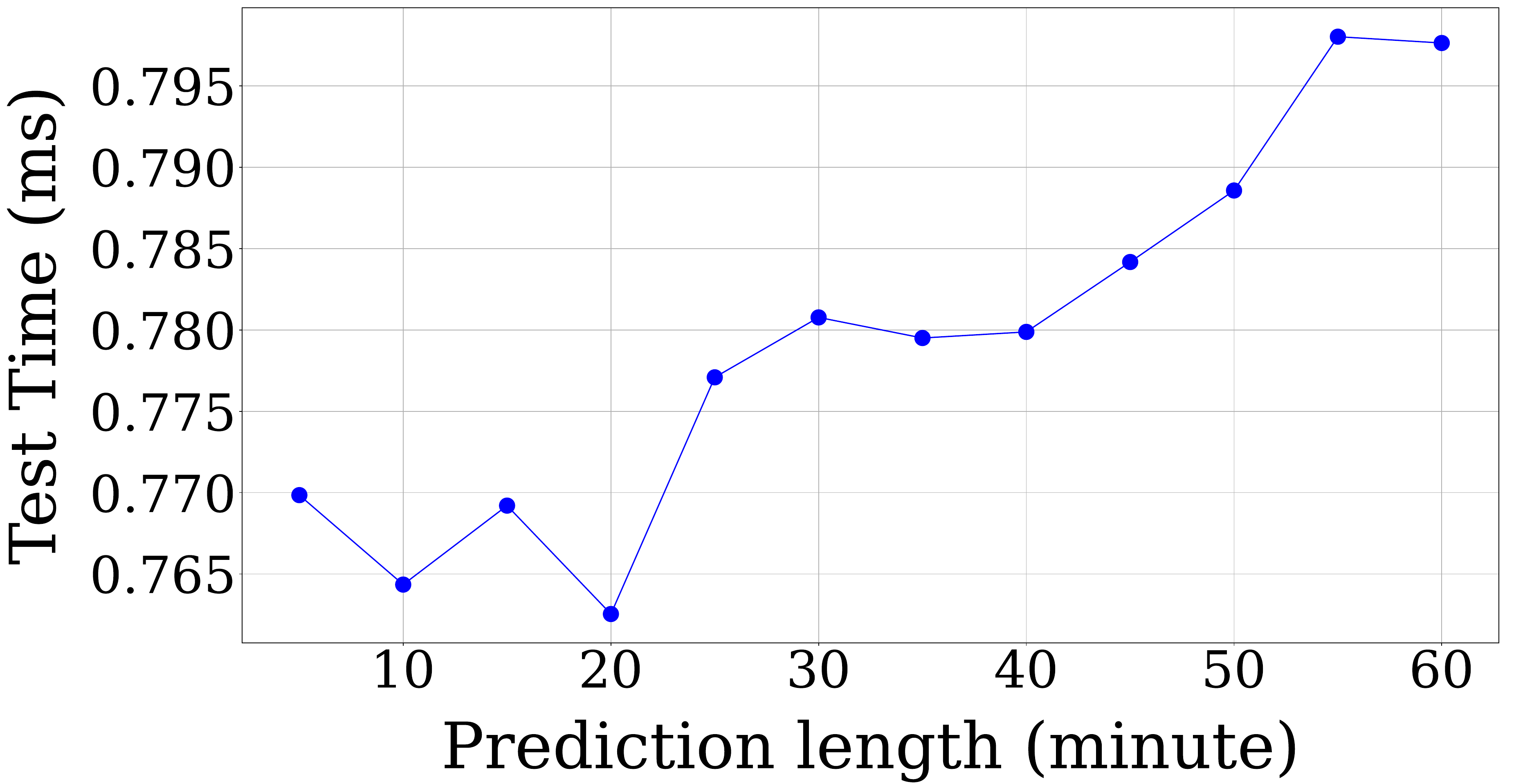}
        \caption{METR-LA dataset.}
        \label{fig:predict_horizons_LA}
    \end{subfigure}%
    \begin{subfigure}{0.45\linewidth}
        \centering
        \includegraphics[width=\textwidth]{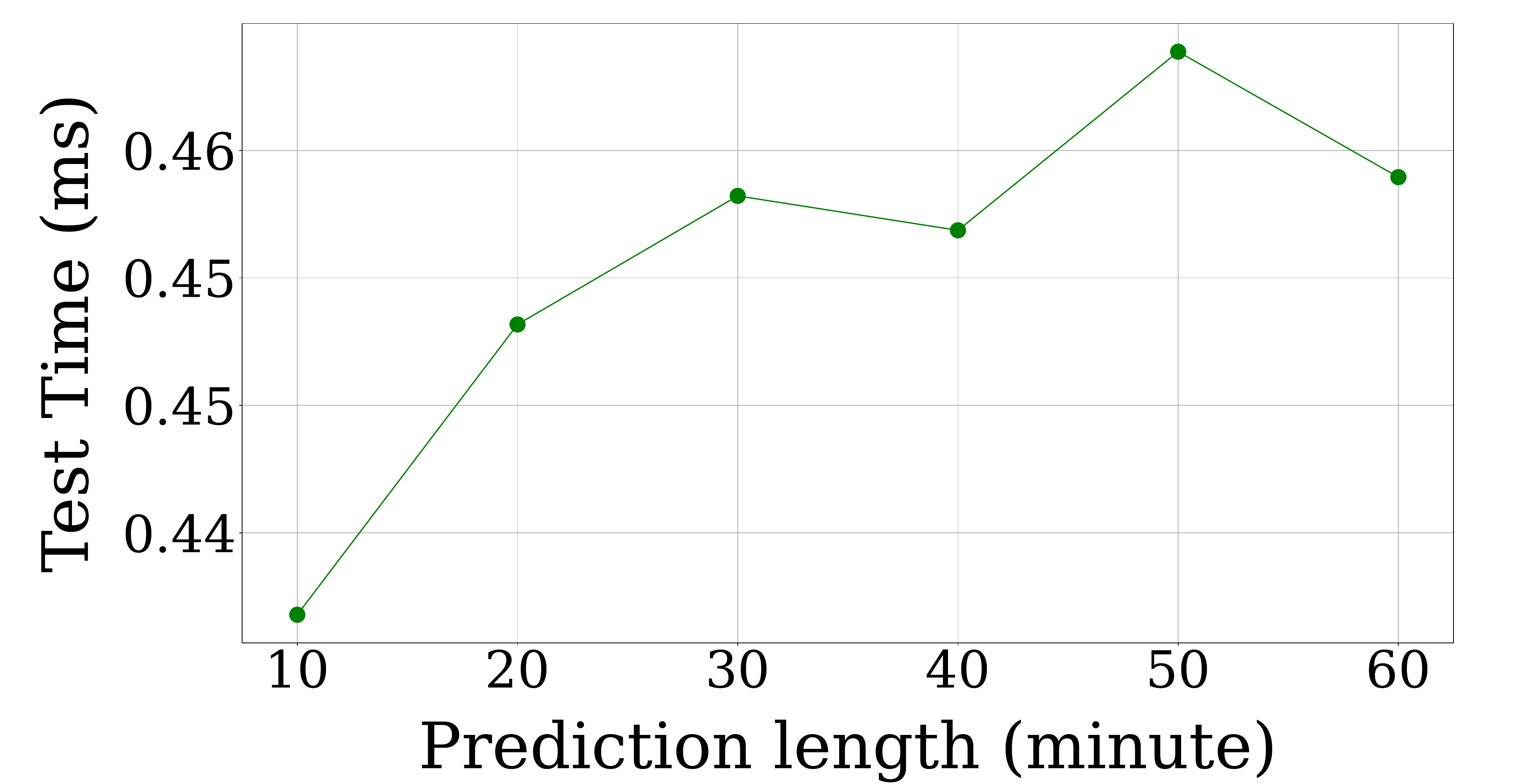}
        \caption{NYCBike1 dataset.}
        \label{fig:predict_horizons_Bike}
    \end{subfigure}
    \caption{Inference time with different prediction lengths $T$.}
    \label{fig:predict_horizons}
    \vspace{-8mm}
\centering
\end{figure}
\subsection{Scalability and Efficiency (RQ5)}\label{sec:scalability}
In this section, we evaluate the scalability and efficiency of MIP to answer RQ5. 
In actuality, MIP needs to be trained and tested on an enlarged dataset with the expansion of urbanization.
Therefore, we record the testing time of one sample in the inference stage with the increasing nodes on the two datasets. As shown in Fig~\ref{fig:node_number}, the testing time increases linearly as the number of nodes increases in urban flow datasets. In usage scenarios, users often request long-term predictions from models, such as navigation software estimating arrival times, which needs to deliver these forecasts quickly, regardless of the travel duration. Therefore, we record the testing time for a single data sample at various future prediction lengths during inference, where the model uses data from the past hour to forecast up to one hour ahead. As shown in Fig.~\ref{fig:predict_horizons_LA}, the testing time on the METR-LA dataset fluctuates when the model makes a prediction within 20 minutes and increases when the prediction length is longer than 25 minutes. As shown in Fig.~\ref{fig:predict_horizons_Bike}, the increase in test time on the NYCBike1 dataset gradually slows down as the prediction length increases and reaches the peak at about 0.465 milliseconds. In a word, in cities with 100 and 200 nodes, MIP can predict flow rates within 0.5 and 1 milliseconds, respectively, for durations up to one hour.

\section{Conclusion}
In this paper, we introduce a new framework named MIP to solve the distribution shift problem in urban flow prediction. MIP stores the most important informative signal during the training process in a memory bank. Then a semantic causality graph structure is established based on the memory bank. Furthermore, the invariant and variant prompts are extracted from the memory bank and we design a spatial-temporal intervention mechanism to create diverse distribution and propose an invariance learning regularization to help the prompt extractor separate the invariant and variant prompts. Extensive experiments on two real-world datasets demonstrate that our method can better handle spatial-temporal distribution shifts than state-of-the-art baselines.

\bibliographystyle{IEEEtran}  
\normalem
\bibliography{ICDEJHY}  

\end{document}